\documentclass{article}

\usepackage{microtype}
\usepackage{graphicx}
\usepackage{subcaption}
\usepackage{booktabs} 

\usepackage{hyperref}
\usepackage{svg}

\usepackage[preprint]{mlconf2026}

\usepackage{amsmath}
\usepackage{amssymb}
\usepackage{mathtools}
\usepackage{amsthm}

\usepackage[capitalize,noabbrev]{cleveref}

\usepackage[textsize=tiny]{todonotes}

\mlconftitlerunning{Critic-Guided Reinforcement Unlearning in Text-to-Image Diffusion}

\begin{document}

\twocolumn[
  \mlconftitle{Critic-Guided Reinforcement Unlearning in Text-to-Image Diffusion}



  \mlconfsetsymbol{equal}{*}

  \begin{mlconfauthorlist}
    \mlconfauthor{Mykola Vysotskyi}{equal,softserve,ucu}
    \mlconfauthor{Zakhar Kohut}{equal,softserve,ucu}
    \mlconfauthor{Mariia Shpir}{softserve}
    \mlconfauthor{Taras Rumezhak}{softserve}
    \mlconfauthor{Volodymyr Karpiv}{softserve}
  \end{mlconfauthorlist}

  \mlconfaffiliation{softserve}{SoftServe, Sadova St, 2D, Lviv, Lviv Oblast Ukraine}
  \mlconfaffiliation{ucu}{Ukrainian Catholic University, Lviv, Ukraine}

  \mlconfcorrespondingauthor{Mykola Vysotskyi}{mvysot@softserveinc.com}
  \mlconfcorrespondingauthor{Zakhar Kohut}{zkohu@softserveinc.com}
  \mlconfcorrespondingauthor{Mariia Shpir}{mshpi@softserveinc.com}
  \mlconfcorrespondingauthor{Taras Rumezhak}{trume@softserveinc.com}
  \mlconfcorrespondingauthor{Volodymyr Karpiv}{vkarpi@softserveinc.com}

  \mlconfkeywords{Machine Learning, Reinforcement Learning, Diffusion Models}

  \vskip 0.3in
]



\printAffiliationsAndNotice{}  

\begin{abstract}
   Machine unlearning in text-to-image diffusion models aims to remove targeted concepts while preserving overall utility. Prior diffusion unlearning methods typically rely on supervised weight edits or global penalties; reinforcement-learning (RL) approaches, while flexible, often optimize sparse end-of-trajectory rewards, yielding high-variance updates and weak credit assignment. We present a general RL framework for diffusion unlearning that treats denoising as a sequential decision process and introduces a timestep-aware critic that predicts expected terminal reward from noisy intermediate states. Concretely, we train a CLIP-based predictor on noisy intermediate states and use it to estimate per-timestep values to compute advantage estimates for policy-gradient updates of the reverse diffusion kernel. Our algorithm is simple to implement, supports off-policy reuse, and plugs into standard text-to-image backbones. Across multiple concepts, the method achieves better or comparable forgetting to strong baselines while maintaining image quality and benign prompt fidelity; ablations show that (i) per-step critics and (ii) noisy-conditioned value estimates are key to stability and effectiveness. We release code and evaluation scripts to facilitate reproducibility and future research on RL-based diffusion unlearning.
\end{abstract}
\section{Introduction}
\label{sec:intro}

Text-to-image diffusion models (DMs) are the backbone of modern image synthesis, yet real deployments increasingly require unlearning—removing specific concepts (styles, identities, unsafe content) without harming overall utility~\cite{ho2020denoisingdiffusionprobabilisticmodels, rombach2022highresolutionimagesynthesislatent}. Existing diffusion unlearning methods typically rely on parameter edits or global penalties and can either over-suppress harmless content or be evaded by adversarial phrasing ~\cite{gandikota2023erasingconceptsdiffusionmodels,gong2024reliable, huang2024recelerreliableconcepterasing,tsai2024ringabellreliableconceptremoval,pham2023circumventingconcepterasuremethods}. In parallel, reinforcement learning (RL) provides a flexible route to optimize non-differentiable or composite objectives by casting denoising as a sequential decision process ~\cite{black2024trainingdiffusionmodelsreinforcement,fan2023dpokreinforcementlearningfinetuning,wallace2023diffusionmodelalignmentusing,clark2024directlyfinetuningdiffusionmodels}. However, most RL-for-diffusion pipelines still depend on end-of-trajectory rewards ~\cite{black2024trainingdiffusionmodelsreinforcement,clark2024directlyfinetuningdiffusionmodels}, which produce high-variance updates and weak credit assignment across many denoising steps. For unlearning—where signals are sparse and boundary precision matters—these limitations are especially challenging ~\cite{feng2025surveygenerativemodelunlearning}.

We propose \textbf{Critic-Guided Reinforcement Unlearning (CGRU)}, an RL framework that treats the reverse diffusion kernel as the policy and augments policy-gradient updates with a per-timestep critic. Concretely, we train a reward predictor on noisy latents to deliver timestep-aware signals; a learned value function over (prompt, latent, timestep) yields advantage estimates that stabilize optimization and sharpen credit assignment. CGRU supports off-policy reuse via importance weighting for sample efficiency ~\cite{10.5555/645531.656005} and integrates with standard text-to-image backbones without architectural changes. Conditioning rewards and values on intermediate latents mitigates two common failure modes in unlearning: (i) over-suppression from global penalties and (ii) instability from sparse terminal rewards. 


We formulate our contributions as follows:

\begin{itemize}
    \item We cast unlearning as RL over the reverse diffusion process and introduce a timestep-aware value function to reduce variance and improve credit assignment ~\cite{schulman2017proximalpolicyoptimizationalgorithms,mohamed2020montecarlogradientestimation, williams1992reinforce}.
    \item We train a CLIP-based classifier on noisy latents to provide per-step signals for advantage-weighted updates, localizing forgetting within the denoising trajectory ~\cite{radford2021learningtransferablevisualmodels}.
    \item We enable off-policy reuse with importance weighting and offer a plug-and-play design—no architectural changes to standard text-to-image backbones ~\cite{10.5555/645531.656005}.
\end{itemize}

\section{Related Works}
\label{sec:related}
\paragraph{Diffusion Models in Brief.} Diffusion models learn to reverse a noise-adding process, gradually denoising latent variables into samples~\cite{ho2020denoisingdiffusionprobabilisticmodels}. Latent-space variants reduce cost while keeping quality~\cite{rombach2022highresolutionimagesynthesislatent}. As DMs are widely deployed, steering them toward desired behaviors has become a key research thread.

\paragraph{Reinforcement-Based Alignment for Diffusion Policies.} Beyond supervised fine-tuning, reinforcement learning (RL) offers a direct way to optimize non-differentiable or composite objectives. A fundamental step is to pose denoising as a sequential decision process and optimize downstream rewards with policy gradients, as in Denoising Diffusion Policy Optimization (DDPO)~\cite{black2024trainingdiffusionmodelsreinforcement}. Subsequent work frames text-to-image fine-tuning with KL-regularized policy gradients (DPOK)~\cite{fan2023dpokreinforcementlearningfinetuning} and adapts Direct Preference Optimization to diffusion likelihoods (Diffusion-DPO)~\cite{wallace2023diffusionmodelalignmentusing}, with large-scale studies reporting multi-objective preference gains~\cite{zhang2024improvinggflownetstexttoimagediffusion}. Other directions include direct rewards backpropagation through the sampler (DRaFT) for differentiable rewards~\cite{clark2024directlyfinetuningdiffusionmodels}, score-based preference optimization~\cite{cai2025dspodirectsemanticpreference}, and extensions to discrete diffusion~\cite{borso2025preferencebasedalignmentdiscretediffusion}. 

Standard policy-gradient tools motivate baselines/critics to reduce variance~\cite{greensmith2004variance}. However, most DM alignment methods use end-of-trajectory rewards, which are sparse and high-variance, highlighting the need for timestep-aware signals. Recent work continues to improve sample efficiency and stability for RL-based diffusion fine-tuning~\cite{gupta2025simpleeffectivereinforcementlearning}.

\paragraph{Selective Forgetting in Generative Models.} Machine unlearning aims to remove specific data or concepts from a trained model for privacy, safety, or compliance. Classic foundations include data deletion and SISA training in discriminative settings~\cite{ginart2019makingaiforgetyou, bourtoule2020machineunlearning}. For diffusion models, concept erasure techniques target styles, identities, or unsafe content while preserving benign capabilities. Representative approaches include ESD~\cite{gandikota2023erasingconceptsdiffusionmodels}, lightweight erasers (Receler)~\cite{huang2024recelerreliableconcepterasing}, closed-form editing (RECE)~\cite{gong2024reliable},  and localized/gated erasure~\cite{lee2025localizedconcepterasuretexttoimage}. 

Robust forgetting remains challenging: erased concepts can often be re-elicited via adversarial prompts~\cite{tsai2024ringabellreliableconceptremoval, pham2023circumventingconcepterasuremethods, beerens2025vulnerabilityconcepterasurediffusion}. Evolving surveys and benchmarks continue to clarify objectives, taxonomies, and metrics in generative unlearning~\cite{feng2025surveygenerativemodelunlearning}; notably, the Holistic Unlearning Benchmark (HUB)~\cite{moon2025holisticunlearningbenchmarkmultifaceted} proposes multi-faceted evaluation across faithfulness, robustness, and alignment dimensions.

\paragraph{Bridging RL Alignment and Unlearning.} Despite fast progress in alignment and unlearning, there are few principled methods that combine them for diffusion models. Some alignment pipelines steer models away from undesirable content using preference-based objectives, DPO-style losses, or KL-regularized updates~\cite{fan2023dpokreinforcementlearningfinetuning, wallace2023diffusionmodelalignmentusing}. Direct Unlearning Optimization (DUO) optimizes paired unsafe/safe preferences to remove targeted concepts while preserving utility~\cite{park2025directunlearningoptimizationrobust}. However, many pipelines still rely on \emph{final-step} rewards or hand-made penalties. In practice, this can yield high-variance credit assignment or, conversely, over-regularization, resulting in overly broad suppression—i.e., the system may unlearn the \emph{core concept} rather than shaping only its boundary.

Our work connects these threads by introducing an RL fine-tuning scheme with a \emph{per-timestep critic}. In contrast to prior erasure methods that edit parameters or embeddings directly~and to DUO’s global preference optimization~\cite{park2025directunlearningoptimizationrobust}—our approach treats the sampler as a sequential policy with timestep-aware feedback. 
\section{Background}
\label{sec:background}

\subsection{Diffusion Models}
\label{subsec:diffusion-models-back}

Diffusion models learn to reverse a noise corruption process \cite{ho2020denoisingdiffusionprobabilisticmodels}. Given clean data $x_0$, the forward process adds Gaussian noise: $q(x_t|x_{t-1}) = \mathcal{N}(x_t; \sqrt{1-\beta_t}x_{t-1}, \beta_t I)$ where $\{\beta_t\}_{t=1}^T$ is a noise schedule. The reverse process learns to denoise by predicting noise: $p_\theta(x_{t-1}|x_t) = \mathcal{N}(x_{t-1}; \mu_\theta(x_t, t), \sigma_t^2 I)$ where $\mu_\theta(x_t, t) = \frac{1}{\sqrt{\alpha_t}}(x_t - \frac{\beta_t}{\sqrt{1-\bar{\alpha}_t}}\epsilon_\theta(x_t, t))$ and $\epsilon_\theta$ is a neural network. The training objective minimizes: $\mathcal{L} = \mathbb{E}_{t,x_0,\epsilon}[\|\epsilon - \epsilon_\theta(\sqrt{\bar{\alpha}_t}x_0 + \sqrt{1-\bar{\alpha}_t}\epsilon, t)\|^2]$.
Here \(\alpha_t := 1-\beta_t\) and \(\bar{\alpha}_t := \prod_{s=1}^{t} \alpha_s\).

\subsection{Reinforcement Learning and Markov Decision Processes}
\label{subsec:rl-mdp-back}

A Markov Decision Process (MDP) is defined as $(S, A, P, R, \gamma)$ where $S$ is the state space, $A$ is the action space, $P(s'|s,a)$ is the transition probability, $R(s,a)$ is the reward function, and $\gamma \in [0,1]$ is the discount factor. The agent follows a policy $\pi(a|s)$ to maximize: $J(\pi) = \mathbb{E}_{\tau \sim \pi}\left[\sum_{t=0}^{\infty} \gamma^t R(s_t, a_t)\right]$ where $\tau = (s_0, a_0, s_1, a_1, \ldots)$ is a trajectory.
In our diffusion-as-RL formulation, we use a \emph{finite-horizon episodic} MDP of length \(T\) with a terminal reward at the end of the denoising trajectory.

\subsection{Connecting Diffusion and Reinforcement Learning}
\label{subsec:diffusion-rl-connection-back}
We cast reverse diffusion as a sequential decision process: the noisy latent and timestep form the state, the sampled previous latent \(x_{t-1}\) is the action, and the reverse kernel acts as the policy. Rewards are assessed on the final sample. This view lets us apply standard policy-gradient tools to diffusion without architectural changes. Full formal definitions and derivations are provided in Appendix~\ref{app:proof}.

\subsection{Policy Gradient Methods for Diffusion}
\label{subsec:policy-gradient-diffusion-back}

Policy gradient methods optimize the objective $J(\theta)$ by leveraging the score function identity \cite{mohamed2020montecarlogradientestimation}. This identity allows us to express the gradient of an expectation as an expectation of the gradient of the log-probability, enabling gradient estimation through sampling.

For the diffusion MDP, by \cite{mohamed2020montecarlogradientestimation}, the score function estimator becomes:
$$
\nabla_\theta J(\theta) = \mathbb{E}_{\tau \sim p_\theta}\left[\sum_{t=1}^{T} \nabla_\theta \log p_\theta(x_{t-1}|x_t) \cdot r(x_0)\right]
$$
Additionally, we adopt importance sampling \cite{10.5555/645531.656005}, reusing off-policy trajectories, to enable multiple optimization steps.

\subsection{Machine Unlearning}
\label{subsec:machine-unlearning}

Machine unlearning addresses the problem of removing specific data points or concepts from a trained model without retraining from scratch. This is particularly important for privacy compliance, data correction, and model adaptation scenarios.

Formally, given a model $f_\theta$ trained on dataset $\mathcal{D}$, unlearning aims to produce a model $f_{\theta'}$ that behaves as if it was trained on $\mathcal{D} \setminus \mathcal{D}_{\text{forget}}$ where $\mathcal{D}_{\text{forget}}$ is the data to be forgotten.

Traditional approaches include: (i) \textbf{Retraining}: Train a new model on $\mathcal{D} \setminus \mathcal{D}_{\text{forget}}$, (ii) \textbf{Fine-tuning}: Continue training on remaining data with regularization, and (iii) \textbf{Approximate unlearning}: Use gradient-based methods to approximate the effect of retraining.

For diffusion models, unlearning becomes particularly challenging due to the iterative nature of the generation process and the complex dependencies between training data and model behavior.
\section{Methodology}
\label{sec:methodology}

We present \textbf{Critic-Guided Reinforcement Unlearning (CGRU)}, a reinforcement learning framework for diffusion unlearning that addresses the limitations of existing approaches. While prior methods rely on end-of-trajectory rewards that yield high-variance updates, CGRU introduces a per-timestep critic that provides dense, informative signals throughout the denoising process.

\subsection{Problem Formulation}

Consider a conditional diffusion model that generates images $x_0$ from prompts $c$ through the reverse process $p_\theta(x_{t-1}|x_t, c)$. For unlearning, we want to optimize the model to minimize the generation of unwanted concepts while preserving overall utility. This can be formulated as maximizing a reward function $r(x_0, c)$ that penalizes unwanted content and rewards desired behavior.

Following the MDP formulation in Appendix~\ref{app:proof}, we treat the reverse diffusion kernel as a policy $\pi_\theta(a_t|s_t) = p_\theta(x_{t-1}|x_t, c)$ where states are $s_t = (c, t, x_t)$ and actions are $a_t = x_{t-1}$. The objective is to maximize:

$$
J(\theta) = \mathbb{E}_{c \sim p(c)} \mathbb{E}_{x_0 \sim p_\theta(x_0|c)}[r(x_0, c)]
$$

\subsection{Critic-Guided Advantage Estimation}

The key innovation of CGRU is the introduction of a per-timestep critic $V_\phi(x_t, c, t)$ that estimates the expected terminal reward from any intermediate state. This critic enables computation of the advantage function $A(s_t) = r(x_0, c) - V_\phi(x_t, c, t)$
at each timestep. In actor–critic terms, $V_\phi$ serves as a learned baseline that reduces variance of REINFORCE-style gradients by centering terminal rewards at each state; the resulting advantage weights provide more informative signals than raw rewards ~\cite{10.5555/645531.656005}.

\subsection{Advantage-Weighted Policy Updates with Importance Sampling}

We prove that incorporating the critic as a baseline does not bias the original objective (due to the fact, that the introduced baseline function is dependent only on the state $(x_t, c, t)$), while significantly reducing variance. The key result is that the gradient can be written as:
$$
\nabla_\theta J(\theta) = \mathbb{E}_{c,\tau \sim p_\theta}\left[\sum_{t=1}^{T} \nabla_\theta \log p_\theta(x_{t-1}|x_t, c) \cdot A(s_t)\right]
$$

In practice, we reuse trajectories within the same outer iteration: after collecting a batch of
$N$ trajectories with policy $\theta_{\mathrm{old}}$, we perform $K$ gradient updates on this batch.
We do not maintain a long-lived replay buffer across many past policies; ``off-policy'' refers to
updating $\theta$ using data sampled from $\theta_{\mathrm{old}}$ via importance weighting \cite{10.5555/645531.656005}.
We monitor importance ratios and stabilize training with gradient clipping; in our settings,
ratios remain well-behaved (see Appendix~\ref{app:additional_results}).

We additionally clip importance ratios to $[\rho_{\min}, \rho_{\max}]$ (PPO-style) \cite{schulman2017proximalpolicyoptimizationalgorithms} for stability; this does not change our conclusions (Appendix~\ref{app:additional_results}).

\subsection{Practical Implementation}

The CGRU framework consists of two main components: critic learning (warm-start + online updates) and policy optimization. We describe the practical implementation details below.

\paragraph{Reward Functions and Datasets.} The choice of reward function and training datasets is crucial for effective unlearning. The reward function $r(x_0, c)$ should penalize the presence of unwanted concepts while preserving overall image quality and prompt adherence. Common approaches include CLIP-based similarity measures to detect concept presence (applied in image space after decoding latents), classifier outputs for specific unwanted content, or other domain-specific metrics depending on the target concept. Importantly, the critic is a standalone module that can be warm-started independently of the diffusion model's fine-tuning process; in our implementation, we continue training it online during policy optimization to reduce distribution shift as the diffusion model evolves.

The training dataset consists of a forgetting dataset containing prompts that directly elicit the concept to be removed. We train only on this forgetting dataset in our experiments.

\paragraph{Critic Training.} The critic $V_\phi(x_t, c, t)$ is trained to predict expected terminal reward from intermediate states through supervised learning on generated trajectories, and it keeps training during diffusion model fine-tuning. Training involves: (1) generating denoising trajectories $(x_{T}, x_{T-1}, \ldots, x_{0})$ for each prompt, (2) computing terminal rewards $r(x_0, c)$, (3) shuffling timesteps to break temporal dependencies, and (4) minimizing the MSE loss:

$$
\mathcal{L}_{\text{critic}} 
= \frac{1}{N} \sum_{i=1}^{N} \sum_{t=1}^{T} 
\left\| V_\phi\!\left(x_t^{(i)}, c^{(i)}, t\right) 
- r\!\left(x_0^{(i)}, c^{(i)}\right) \right\|^2
$$

In our implementation, $V_\phi$ is instantiated with a CLIP-based vision classifier: each intermediate state $x_t$ (a noisy latent) is decoded to RGB space using the frozen VAE decoder, and the resulting image is fed to the CLIP vision encoder with a lightweight classification head.

The critic uses sinusoidal timestep embeddings and FiLM layers~\cite{perez2017filmvisualreasoninggeneral} for temporal awareness, enabling meaningful value estimates across denoising timesteps. Details are in Appendix~\ref{app:film}.

\paragraph{Policy Optimization.} Policy optimization uses advantage-weighted gradients: (1) compute advantages $A_t = r(x_0, c) - V_\phi(x_t, c, t)$ at every timestep, (2) shuffle timesteps to break temporal dependencies (as in critic training), (3) apply importance weighting for off-policy data, and (4) update the policy with stability controls. To ensure stable off-policy updates, we employ gradient clipping, following DDPO~\cite{black2024trainingdiffusionmodelsreinforcement}; these mechanisms constrain the policy update magnitude and prevent excessively large steps that could destabilize training. 

All details on objective derivations and proofs are available in Appendix~\ref{app:proof} and complete algorithms are in Appendix~\ref{app:algorithms}.

\section{Experiments}
\label{sec:experiments}

We present experimental validation of our CGRU framework through comparative studies with existing baselines. Our experiments demonstrate the practical benefits of our per-timestep critic approach across different settings and objectives.

\subsection{Experimental Setup}

All experiments were conducted using Stable Diffusion 1.5 as the base diffusion model, with LoRA (Low-Rank Adaptation) applied for memory efficiency during fine-tuning. Training was performed using gradient accumulation with 2 updates per epoch, requiring approximately 35 GB of VRAM on a single H100 GPU. Each training run took approximately 2--5 hours (100--250 epochs), depending on the target class and its convergence speed. Detailed hyperparameters and training configurations are provided in the Appendix~\ref{app:training}.
We aggregate results over 3 random seeds.

\subsection{Comparison with DDPO}
We compare CGRU against DDPO~\cite{black2024trainingdiffusionmodelsreinforcement} using the LAION Aesthetic Score \cite{schuhmann2022laion} reward function - aesthetics predictor, built as a linear layer applied to CLIP embeddings. Although DDPO is not a specialized unlearning method, it serves as an ablation baseline to validate our theoretical claims regarding the superiority of dense, timestep-aware signals over sparse terminal rewards. Both methods share the same base model, reward, dataset, etc. The key difference is the learning algorithm itself: CGRU augments the update with a per-timestep critic (enabling dense advantages and trajectory reuse), whereas DDPO follows the standard on-policy baseline with terminal rewards.

\begin{figure}[h]
    \centering
    \includegraphics[width=\columnwidth]{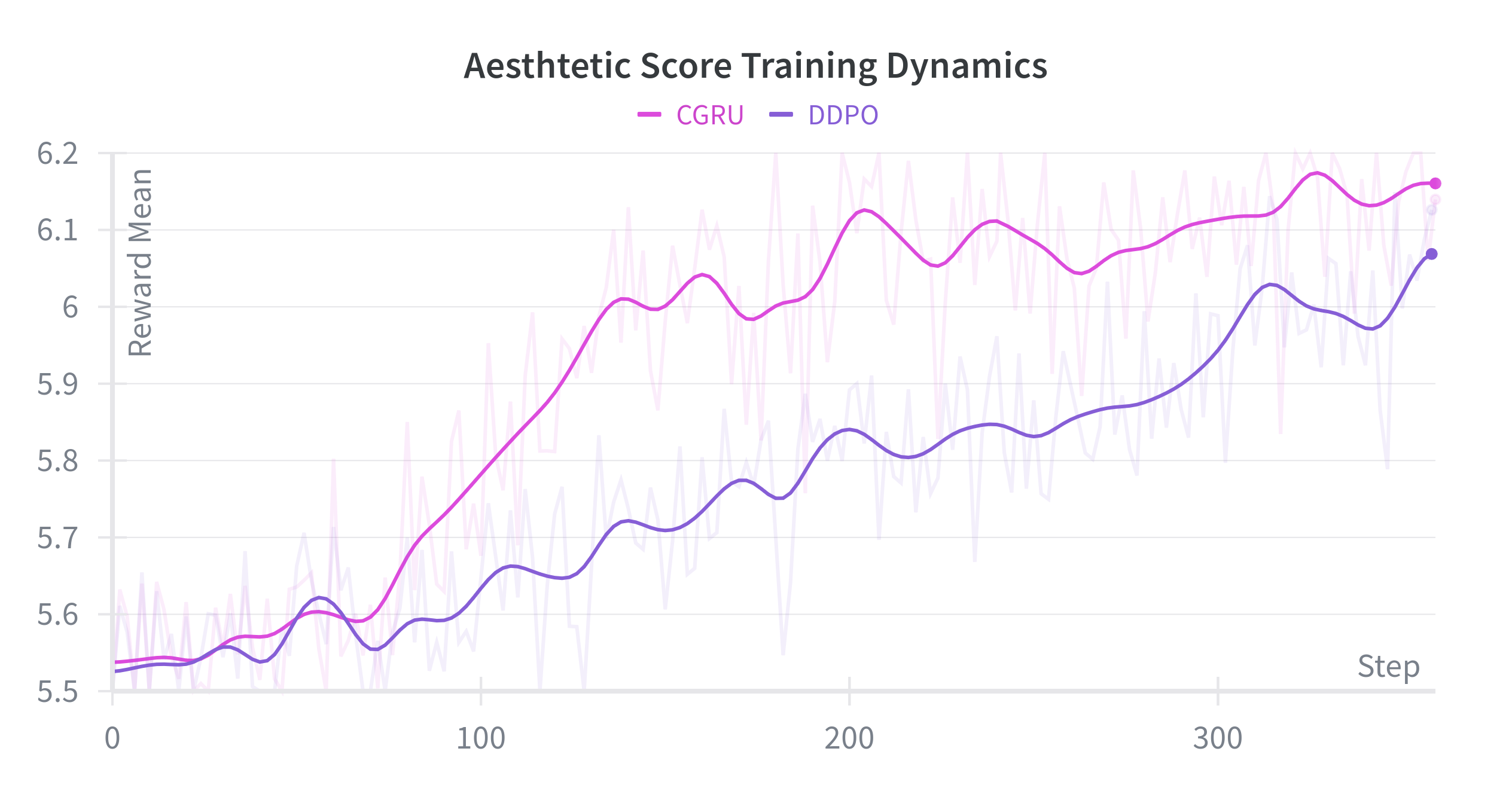}
    \caption{Mean Aesthetic Score reward during training for CGRU and DDPO. CGRU shows superior performance with faster convergence and higher final rewards.}
    \label{fig:aesthetic_comparison}
\end{figure}

Figure~\ref{fig:aesthetic_comparison} shows the mean Aesthetic Score reward over the course of training for both methods. The results demonstrate that CGRU consistently outperforms DDPO, achieving higher final rewards and showing faster convergence throughout the optimization process.

The improved performance of CGRU validates our theoretical analysis, suggesting that the per-timestep critic provides more effective optimization signals compared to sparse, end-of-trajectory rewards. Additionally, we observe that CGRU exhibits a lower gradient variance during training (see Appendix~\ref{subsec:grad_var}), which confirms the stability benefits of our approach. This result demonstrates the practical viability of CGRU for the optimization of RL-based diffusion models.

\subsection{Concept Removal}

To validate CGRU's effectiveness for machine unlearning, we conducted experiments following established evaluation protocols for concept removal in diffusion models~\cite{zhang2024unlearncanvas}. Our experimental pipeline encompasses 20 distinct object classes, enabling systematic assessment of our method's ability to suppress specific concepts while maintaining overall generation quality.

We trained a CLIP-based classifier to serve as our reward function, utilizing the \textit{"openai/clip-vit-base-patch32"} model with classification head. The classifier outputs probabilities for the 20 object classes. We used the complement probability of the target class, scaled in the 0-10 range, as a reward signal.

With the reward function established, we warm-start the per-timestep critic network using the same CLIP model for consistency and continue updating it online during diffusion model fine-tuning.

We used curated subsets from the referenced dataset~\cite{kumari2023ablatingconceptstexttoimagediffusion}, partitioning the data into three distinct components: (1) classifier training data for reward function development, (2) critic training trajectories for value function estimation, and (3) unlearning prompts for policy optimization.

After that we applied CGRU to "unlearn" each of the 20 object classes. 



\subsection{NSFW Content Removal.}
\label{subsec:nsfw}
To test whether CGRU extends beyond object-level erasure to safety-critical concept removal, we apply it to suppress \emph{nudity/NSFW} generations and evaluate on a standardized prompt suite, specifically - I2P benchmark \cite{schramowski2023safelatentdiffusionmitigating}. This setting is challenging because the target concept is highly abstract and can be elicited by diverse prompts.

We use a public binary NSFW detector~\cite{falcons_nsfw} (a fine-tuned vision transformer) as the reward function. Concretely, given an image \(x\), the detector outputs a probability of NSFW content \(p_{\text{nsfw}}(x)\); we set the reward as \(r(x)=1-p_{\text{nsfw}}(x)\), so higher reward corresponds to safer generations. We warm-start a per-timestep critic with the same backbone architecture and FiLM conditioning as in our CGRU setup for concept removal and keep updating it online to predict the expected terminal reward from intermediate noisy latents.

To obtain a diverse distribution of explicit and borderline content, we draw training images from the \textit{NSFW Detection Dataset} by DeepGHS~\cite{deepghs_nsfw_detect}, which contains five categories (\texttt{drawing}, \texttt{hentai}, \texttt{neutral}, \texttt{porn}, \texttt{sexy}). We train the critic on a random 10\% split and perform policy optimization on a disjoint 20\% split. Since RL fine-tuning requires text prompts, we convert images into prompts by re-captioning each image using LLaVA-1.5-7B~\cite{liu2023visualinstructiontuning}. We use a safety-aware instruction (Appendix~\ref{app:llava_safety_prompt}) that explicitly requests a literal description and asks the model to include NSFW or suggestive elements when present; this reduces prompt ambiguity and makes the unlearning objective better aligned with the safety concept being removed. We further assess robustness to adversarial prompt attacks on the UnlearnDiff nudity benchmark and its accompanying UnlearnDiffAtk procedure (adversarial prompt-rewriting attack)~\cite{zhang2024generatenotsafetydrivenunlearned}. \Cref{tab:unlearndiffatk_nudity} reports the Attack Success Rate (ASR) before and after applying UnlearnDiffAtk; CGRU achieves 0.00 pre-ASR and 9.67 post-ASR.

\begin{table}[t]
    \caption{Attack Success Rate (ASR) before and after UnlearnDiffAtk~\cite{zhang2024generatenotsafetydrivenunlearned} on UnlearnDiff nudity prompts (lower is better).}
    \label{tab:unlearndiffatk_nudity}
    \centering
    \small
    \begin{tabular}{l|cc}
        \toprule
        Method & Pre-ASR & Post-ASR \\
        \midrule
        FMN & 88.03 & 97.89 \\
        SPM & 54.93 & 91.55 \\
        SAFREE & 26.06 & 85.59 \\
        UCE & 21.83 & 79.58 \\
        ESD & 20.42 & 76.05 \\
        RECE & 13.38 & 75.42 \\
        MACE & 9.10 & 74.57 \\
        AdvUn & 7.75 & 21.13 \\
        SalUn & 1.41 & 11.27 \\
        SHS & 0.00 & 7.04 \\
        Erasediff & 0.00 & 2.11 \\
        CGRU (ours) & \textbf{0.00} & \textbf{9.67} \\
        \bottomrule
    \end{tabular}
\end{table}

\subsection{Ablation: Impact of Timestep Awareness}
\label{subsec:ablation_film}

To validate the importance of our timestep-aware design, we compare the FiLM-conditioned critic against a standard CLIP-based critic without timestep conditioning. Both models are evaluated on their ability to predict the final reward from intermediate noisy latents. Since the reward derives from a 20-class concept classifier, we frame this as a classification task over reward levels; test samples are drawn uniformly across timesteps to cover the full noise spectrum.

The timestep-aware critic achieves 43.25\% accuracy and 58.52\% macro precision, compared to 29.75\% and 34.47\% for the baseline (see Figure~\ref{fig:ablation_metrics} in Appendix~\ref{app:film}). This substantial improvement confirms that FiLM-based timestep conditioning is essential for accurate value estimation from noisy states, particularly in the earlier stages of diffusion where the signal-to-noise ratio is low.
   
\section{Evaluation}
\label{sec:evaluation}

We evaluate CGRU on (i) object concept removal using the UnlearnCanvas benchmark~\cite{zhang2024unlearncanvas} and (ii) safety-critical nudity suppression using the Inappropriate-to-Proper (I2P) benchmark from Safe Latent Diffusion~\cite{schramowski2023safelatentdiffusionmitigating}. We report both quantitative metrics and qualitative evidence to characterize the erasure--utility trade-off.

\subsection{Quantitative Results}

We first evaluate CGRU on UnlearnCanvas~\cite{zhang2024unlearncanvas} using the ViT classifier provided by the benchmark. We focus on three metrics that capture different aspects of unlearning: \textbf{Unlearning Accuracy (UA)} measures the proportion of samples generated from prompts containing the target concept that are \textbf{not} correctly classified by the concept detector. \textbf{In-domain Retain Accuracy (IRA)} quantifies the model's ability to correctly generate and classify samples containing other concepts from the same domain, ensuring that unlearning does not severely harm unrelated capabilities. We also measure image quality through \textbf{Fr\'echet Inception Distance (FID)} \cite{heusel2018gans} to ensure that unlearning does not  compromise the visual fidelity of generated images.

\begin{table*}[h]
\centering
\caption{\textbf{Comparison of CGRU with state-of-the-art unlearning methods on object removal tasks.} Best results are highlighted in bold, second-best are underlined. CGRU achieves strong unlearning accuracy while maintaining competitive retain accuracy.}
\label{tab:unlearn_overview}
\small
\begin{tabular}{l|cc|c|c}
\toprule
\textbf{Method} & \textbf{UA} ($\uparrow$) & \textbf{IRA} ($\uparrow$) & \textbf{Average} ($\uparrow$) & \textbf{FID} ($\downarrow$) \\
\midrule
ESD \cite{gandikota2023erasingconceptsdiffusionmodels}  & $92.15\%$ & $55.78\%$ & $73.97\%$ & $65.55$ \\
FMN \cite{zhang2024unlearncanvas}  & $45.64\%$ & $90.63\%$ & $68.14\%$ & $131.37$ \\
UCE \cite{gandikota2024unifiedconcepteditingdiffusion}  & $\underline{94.31}\%$ & $39.35\%$ & $66.83\%$ & $182.01$ \\
CA \cite{kumari2023ablatingconceptstexttoimagediffusion}  & $46.67\%$ & $90.11\%$ & $68.39\%$ & $\textbf{54.21}$ \\
SalUn \cite{fan2024salunempoweringmachineunlearning} & $86.91\%$ & $\mathbf{96.35}\%$ & $\mathbf{91.63}\%$ & $61.05$ \\
SEOT \cite{li2024wantdontimagecontent} & $23.25\%$ & $\underline{95.57}\%$ & $59.41\%$ & $62.38$ \\
SPM \cite{lyu2024onedimensionaladapterruleall} & $71.25\%$ & $90.79\%$ & $81.02\%$ & $\underline{59.79}$ \\
EDiff \cite{wu2024erasingundesirableinfluencediffusion} & $86.67\%$ & $94.03\%$ & $\underline{90.35\%}$ & $81.42$ \\
SHS \cite{wu2024scissorhandsscrubdatainfluence} & $80.73\%$ & $81.15\%$ & $80.94\%$ & $119.34$ \\
SAeUron \cite{cywiński2025saeuroninterpretableconceptunlearning} & $78.82\%$ & $95.47\%$ & $87.15\%$ & $62.15$ \\
\textbf{CGRU} & $\mathbf{95.37}\%$ & $81.64\%$ & $88.50\%$ & $68.43$ \\
\bottomrule
\end{tabular}

\end{table*}

Table~\ref{tab:unlearn_overview} summarizes object unlearning results. Baseline numbers are taken from the recent consolidated benchmark in~\cite{cywiński2025saeuroninterpretableconceptunlearning}; CGRU is evaluated with the same UnlearnCanvas protocol. CGRU attains the highest UA (95.37\%), while trading off IRA.

\begin{table*}[h]
\centering
\caption{\textbf{Nudity unlearning evaluation on the I2P benchmark.} CLIPScore is calculated for the NSFW experiment model only. The best result for each metric is highlighted in bold.}
\label{tab:unlearn_nudity}
\resizebox{\textwidth}{!}{%
\begin{tabular}{l|ccccccccc|c}
\toprule[1pt]
\midrule
\textbf{Method} & \textbf{Armpits} & \textbf{Belly} & \textbf{Buttocks} & \textbf{Feet} & \textbf{Breasts (F)} & \textbf{Genitalia (F)} & \textbf{Breasts (M)} & \textbf{Genitalia (M)} & \textbf{Total} & \textbf{CLIPScore} ($\uparrow$) \\
\midrule
FMN \cite{zhang2024unlearncanvas} & 43 & 117 & 12 & 59 & 155 & 17 & 19 & 2 & 424 & 30.39 \\
CA \cite{kumari2023ablatingconceptstexttoimagediffusion} & 153 & 180 & 45 & 66 & 298 & 22 & 67 & 7 & 838 & \textbf{31.37} \\
AdvUn \cite{zhang2024defensiveunlearningadversarialtraining} & 8 & \textbf{0} & \textbf{0} & 13 & \textbf{1} & 1 & \textbf{0} & \textbf{0} & 28 & 28.14 \\
Receler \cite{huang2024recelerreliableconcepterasing} & 48 & 32 & 3 & 35 & 20 & \textbf{0} & 17 & 5 & 160 & 30.49 \\
MACE \cite{lu2024macemassconcepterasure} & 17 & 19 & 2 & 39 & 16 & \textbf{0} & 9 & 7 & 111 & 29.41 \\
CPE \cite{lee2025conceptpinpointerasertexttoimage} & 10 & 8 & 2 & 8 & 6 & 1 & 3 & 2 & 40 & 31.19 \\
UCE \cite{gandikota2024unifiedconcepteditingdiffusion} & 29 & 62 & 7 & 29 & 35 & 5 & 11 & 4 & 182 & 30.85 \\
SLD-M \cite{schramowski2023safelatentdiffusionmitigating} & 47 & 72 & 3 & 21 & 39 & 1 & 26 & 3 & 212 & 30.90 \\
ESD \cite{gandikota2023erasingconceptsdiffusionmodels} & 32 & 30 & 2 & 19 & 27 & 3 & 8 & 2 & 123 & 30.21 \\
SAeUron \cite{cywiński2025saeuroninterpretableconceptunlearning} & 7 & 1 & 3 & \textbf{2} & 4 & \textbf{0} & \textbf{0} & 1 & \textbf{18} & 30.89 \\
\textbf{CGRU} & \textbf{2} & 6 & 7 & 6 & 17 & \textbf{0} & \textbf{0} & 2 & 40 & 30.27 \\
\midrule
SD v1.4 & 148 & 170 & 29 & 63 & 266 & 18 & 42 & 7 & 743 & 31.34 \\
SD v2.1 & 105 & 159 & 17 & 60 & 177 & 9 & 57 & 2 & 586 & 31.53 \\
\midrule
\bottomrule[1pt]
\end{tabular}%
}
\vspace*{-1em}
\end{table*}

\begin{figure}[h]
    \centering
    \includegraphics[width=\columnwidth]{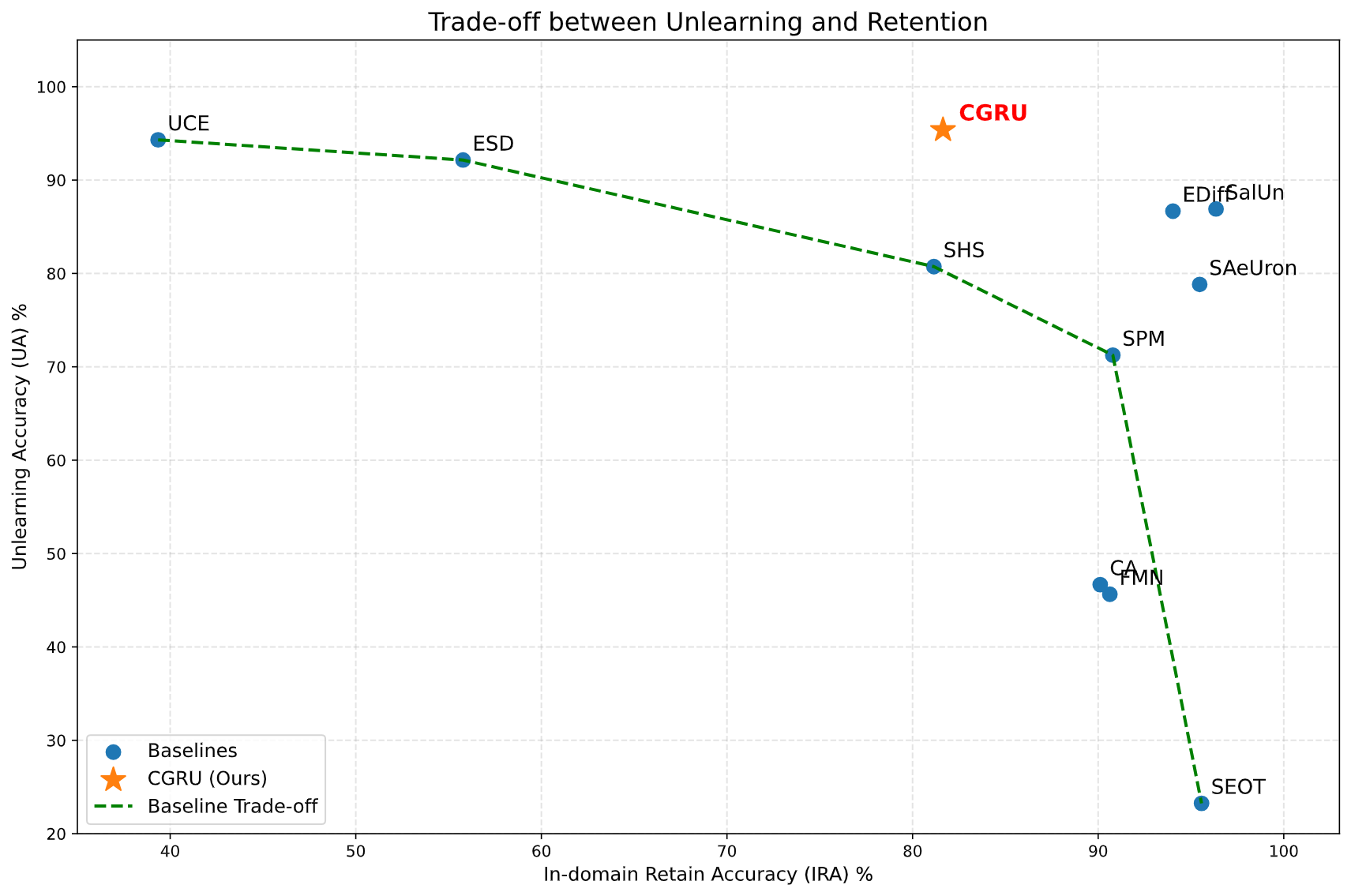}
    \caption{\textbf{UA--IRA trade-off (Pareto frontier).} Each point is a method from Table~\ref{tab:unlearn_overview}. CGRU achieves the strongest erasure (UA) and lies above the baseline trend for high-erasure regimes, highlighting it as a Pareto-competitive choice when prioritizing concept removal.}
    \label{fig:pareto_tradeoff}
\end{figure}
Across methods, there is a clear tension between suppressing the target concept (high UA) and preserving in-domain capability for related concepts (high IRA), consistent with prior observations in diffusion unlearning. Table~\ref{tab:unlearn_overview} shows that CGRU achieves the strongest UA (95.37\%) while maintaining a solid IRA (81.64\%), yielding the third-best overall Average score (88.50\%). While retention-optimized baselines such as SalUn and SAeUron attain higher IRA, they do so with weaker UA; conversely, other high-UA erasure methods (e.g., ESD/UCE) sacrifice substantially more IRA. Figure~\ref{fig:pareto_tradeoff} highlights this regime: CGRU sits above the baseline UA--IRA trend among common erasure methods (ESD/UCE/SHS/SPM/SEOT), making it the best choice when UA must be prioritized. Second, we evaluate nudity suppression on I2P using NudeNet~\cite{nudenet_cite} with a 0.6 confidence threshold (following~\cite{schramowski2023safelatentdiffusionmitigating}); detections are aggregated per category and summarized as a total. On I2P, CGRU reduces the total nudity detections to 40 (Table~\ref{tab:unlearn_nudity}), substantially improving over concept erasure baselines such as ESD and UCE, and approaching the strongest reported methods such as SAeUron. Importantly, we also report CLIPScore to monitor prompt adherence: while aggressive safety unlearning can reduce alignment to the input text, CGRU maintains acceptable semantic fidelity. We note that CGRU's CLIPScore (30.27) is lower than retention-focused baselines; this reflects a known trade-off where aggressive nudity suppression can reduce prompt adherence, particularly when prompts themselves contain suggestive language. Balancing erasure with explicit alignment preservation (e.g., via multi-objective rewards) is a promising direction for future work. Overall, this result suggests that the critic-guided RL framework extends beyond object-level erasure to a more abstract, safety-critical concept (nudity), using an evaluation protocol that is independent of the training reward.

\subsection{Qualitative Results}

\begin{figure*}[t]
    \centering
    \includegraphics[width=0.85\textwidth]{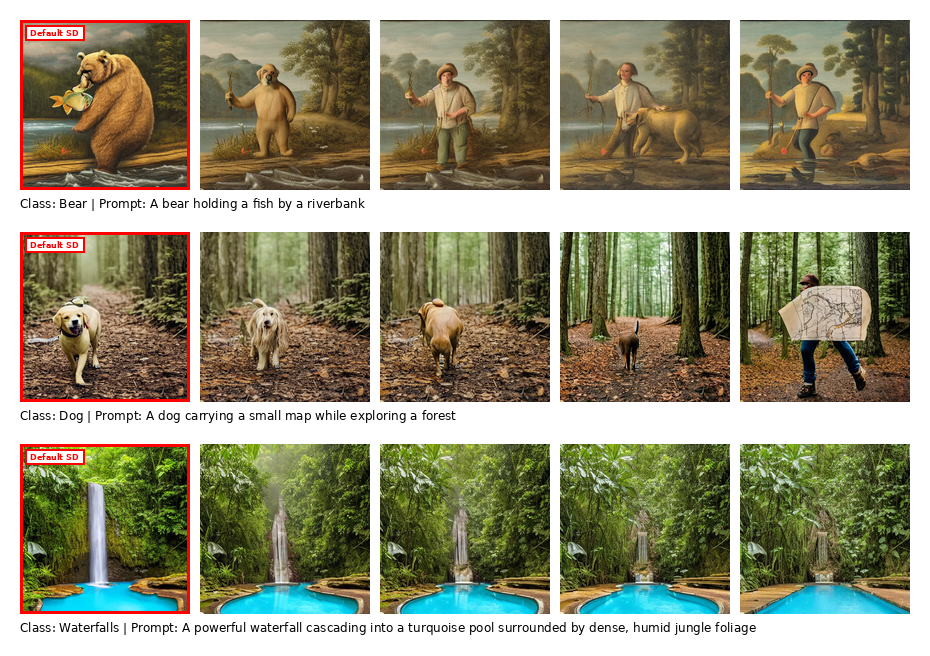}
    \caption{Progression of concept forgetting during training for CGRU. Images are sampled from different training checkpoints (default, early, mid, late), illustrating the gradual suppression of the target concept.}
    \label{fig:forgetting_progression}
\end{figure*}

Qualitatively, we visualize how concept evidence decays as RL fine-tuning progresses. Figure~\ref{fig:forgetting_progression} shows generations from intermediate checkpoints for a representative prompt under CGRU.
Beyond simply removing a target concept, we observe that the model often ``fills in'' the scene in a visually plausible way: the removed object may be replaced by a different semantically compatible subject (e.g., another object or a human figure), blended into the surrounding environment, or expressed only through non-diagnostic background elements. In other words, forgetting typically manifests as a natural re-composition of the image rather than a degenerate collapse into noise or low-quality artifacts.
To avoid overloading the main paper with large image grids, we defer additional qualitative progressions (Figures~\ref{fig:progression_appendix_1}--\ref{fig:progression_appendix_2}), per-class UA/IRA (Table~\ref{tab:per_class_metrics}), and per-class misclassification matrices (Figure~\ref{fig:misclassification_matrices}) to Appendix~\ref{app:additional_results}. We also report a gradient-variance analysis supporting the stability claim (Appendix~\ref{subsec:grad_var}) and details on the FID/CLIPScore calculation in the Appendix~\ref{app:evalmetr}.

\section{Limitations \& Future Work}
\label{sec:discussion}

While our experiments demonstrate effectiveness on both object concept removal (UnlearnCanvas) and safety-critical nudity suppression (I2P), the generalizability of CGRU beyond these settings remains to be investigated.

\paragraph{Limitations.} CGRU relies on task-specific automated reward signals (e.g., concept detectors or NSFW classifiers) and a critic that must learn to predict them from noisy latents; performance is therefore bounded by the fidelity and coverage of these rewards. Moreover, RL fine-tuning is computationally expensive, which limits rapid prototyping and large-scale deployment. Unlike some parameter editing methods that offer stronger formal guarantees, our reinforcement learning approach provides limited guarantees about unlearning completeness. The success relies on learned components subject to approximation errors. Additionally, the effectiveness of CGRU depends on the quality of the reward model; a poorly specified detector may lead to reward hacking (e.g., generating images that fool the detector without true concept removal). We mitigate this by following established benchmarks and using publicly available detector architectures, but ensemble rewards or human evaluation could provide additional robustness. As with any output-level optimization, CGRU may re-route generation pathways rather than fully erase latent concept representations. The qualitative evidence (Figure~\ref{fig:forgetting_progression}) suggests the model redistributes visual content in a semantically plausible way rather than collapsing to noise, which is desirable for image quality but leaves open the question of whether deeper concept removal occurs. More generally, unlearning may hold under the evaluated prompt distribution while still allowing regrowth under distribution shift or adaptive prompting; while we include an adversarial prompt evaluation for nudity (Table~\ref{tab:unlearndiffatk_nudity}), robustness to stronger attacks remains open. Finally, we do not include an explicit retention loss or retain-set training; retention is an emergent property of localized parameter updates and the base model prior, and can degrade under distribution shift.


\paragraph{Future work.}
Promising directions include improving robustness under prompt variation and adaptive attacks, and developing more principled multi-objective formulations that explicitly trade off erasure and retention. It will also be valuable to extend CGRU beyond object removal and nudity suppression (e.g., to styles, identities, or other safety domains). Evaluating on newer diffusion backbones (e.g., SDXL, SD3) is another important step; we focus on Stable Diffusion 1.5 primarily due to the computational cost of RL training and because it remains the common evaluation standard for state-of-the-art baselines. Finally, we believe it is promising to study where and how unlearning manifests inside the model using interpretability and parameter/representation analysis methods (e.g., weight/activation attribution, probing, sparse autoencoders, or related mechanistic analyses), with the goal of making unlearning more targeted and easier to diagnose.

\section{Conclusion}
\label{sec:conclusion}

We presented Critic-Guided Reinforcement Unlearning (CGRU), an RL framework for diffusion unlearning that treats reverse denoising as a sequential decision process and augments policy-gradient updates with a timestep-aware critic. By conditioning value estimation on intermediate noisy latents, CGRU provides dense learning signals across the trajectory, improving credit assignment and stabilizing training compared to sparse terminal-reward optimization. Empirically, CGRU achieves strong concept suppression on UnlearnCanvas, reaching 95.37\% unlearning accuracy while maintaining reasonable retention and image quality, and extends to safety-critical nudity suppression on I2P. We complement aggregate benchmark numbers with per-class diagnostics and misclassification analyses in the appendix, and include supporting ablations and stability evidence. Our code and evaluation scripts are available at~\cite{cgru-2026} to facilitate reproducibility and follow-up research on RL-based diffusion unlearning.

\clearpage
\bibliography{main}
\bibliographystyle{mlconf2026}

\newpage
\appendix
\onecolumn

\section{Proof of Advantage-Weighted Policy Gradient for Diffusion}
\label{app:proof}

\subsection{Setting and notation.}
Let $x_0$ denote the final sample generated from a conditional diffusion model given context $c$. Sampling proceeds by drawing $x_T \sim \mathcal{N}(0,I)$ and iteratively sampling $x_{t-1} \sim p_\theta(x_{t-1}|x_t, c)$ to obtain trajectory $\tau = (x_{T}, x_{T-1}, \ldots, x_{0})$ with terminal reward $r(x_0, c)$. Viewing denoising as a multi-step MDP with states $s_t=(x_t,c,t)$ and actions $a_t=x_{t-1}$, the policy is $\pi_\theta(a_t|s_t) \equiv p_\theta(x_{t-1}|x_t, c)$ and rewards are zero at all $t>0$ and $r(x_0, c)$ at $t=0$. The MDP can be written as
\begin{align}
s_t &= (x_t, c, t) \\
a_t &= x_{t-1} \\
\pi_\theta(a_t|s_t) &= p_\theta(x_{t-1}|x_t, c) \\
P(s_{t+1}|s_t, a_t) &= \delta_{x_{t-1}} \otimes \delta_{t-1} \\
R(s_t) &= \begin{cases} r(x_0, c) & \text{if } t = 0 \\ 0 & \text{otherwise.} \end{cases}
\end{align}
Here the context \(c\) is fixed for the entire episode and therefore omitted from the transition kernel notation.
The initial state distribution is $\rho_0(s_0) = p(x_T) \otimes \delta_T$ with $x_T \sim \mathcal{N}(0,I)$. The objective is the expected terminal reward under the induced sample distribution:
\begin{equation}
\label{eq:j-ddpo}
J(\theta) = \mathbb{E}_{c \sim p(c)} \mathbb{E}_{x_0 \sim p_\theta(x_0|c)}[r(x_0, c)].
\end{equation}

\subsection{Objective and the DDPO score-function gradient}
\label{sec:ddpo-objective}

In DDPO \cite{black2024trainingdiffusionmodelsreinforcement}, the authors take advantage of the stepwise likelihoods $p_\theta(x_{t-1}|x_t, c)$ and apply the score-function (likelihood-ratio) identity at each denoising step:
\begin{align}
\label{eq:grad-ddpo}
\nabla_\theta J(\theta) = \mathbb{E}_{c,\tau \sim p_\theta}\left[\sum_{t=1}^{T}\nabla_\theta \log p_\theta(x_{t-1}|x_t, c) \cdot r(x_0, c)\right]
\end{align}

\subsection{What is the critic in our setting?}
\label{subsubsec:critic}

We introduce a per-timestep critic (a value function) $V_\phi(x_t, c, t)$ that predicts the expected terminal reward conditioned on the current state and context:
\begin{equation}
\label{eq:value-def}
V_\phi(x_t, c, t) \approx \mathbb{E}_{\tau_{0:t-1} \sim p_\theta(\cdot | x_t, c)}[r(x_0, c) | x_t, c, t]
\end{equation}

i.e., the expected downstream reward from the partial reverse trajectory starting at $(x_t, c, t)$. This critic serves as a baseline to reduce variance of the gradient. It can be trained by Monte Carlo regression against the realized terminal reward:
\begin{equation}
\label{eq:value-regression}
\min_{\phi} \mathbb{E}_{c,\tau \sim p_\theta}\left[\sum_{t=1}^{T} (r(x_0, c) - V_\phi(x_t, c, t))^2\right]
\end{equation}

optionally with replay and/or target networks. Intuitively, $V_\phi$ spreads the sparse terminal reward back over intermediate latent states, stabilizing policy updates.

\subsection{Baseline-augmented (advantage) gradient}
\label{subsubsec:baseline}

Let's start from the score-function identity \cite{mohamed2020montecarlogradientestimation}.
For any integrable function $f(\tau)$ and policy $p_\theta(\tau)$,
\[
\nabla_\theta \mathbb{E}_{\tau \sim p_\theta}[f(\tau)]
= \mathbb{E}_{\tau \sim p_\theta}[f(\tau) \nabla_\theta \log p_\theta(\tau)]
\]

Under our factorization $p_\theta(\tau|c)=\prod_{t=1}^T p_\theta(x_{t-1}|x_t, c)$, we write
\[
\nabla_\theta \log p_\theta(\tau|c)
= \sum_{t=1}^{T} \nabla_\theta \log p_\theta(x_{t-1}|x_t, c)
\]

Plugging $f(\tau)=r(x_0, c)$ and then averaging over $c \sim p(c)$ gives Eq.~\eqref{eq:grad-ddpo}.

\bigskip
\textbf{The next step is introduction of the critic.} We will integrate it into the loss, prove that the change does not bias the original objective and the impact of such changes.
Fix an arbitrary baseline function $b(x_t, c, t)$ that depends only on the state $x_t$ (it may also depend
on $c$ and $t$, but it must not depend on the sampled action $x_{t-1}$). 
Start from the gradient expression and add zero in the form ``$B-B$'' where $B$ is the baseline-weighted term:
\begin{align}
\nonumber \nabla_\theta J(\theta) = \mathbb{E}_{c,\tau}\left[\sum_{t=1}^{T} \nabla_\theta\log p_\theta(x_{t-1}|x_t, c) \cdot r(x_0, c)\right] + \mathbb{E}_{c,\tau}\left[\sum_{t=1}^{T}
\nabla_\theta\log p_\theta(x_{t-1}|x_t, c) \cdot b(x_t, c, t)\right] 
\\\nonumber - \underbrace{\mathbb{E}_{c,\tau}\left[\sum_{t=1}^{T}
\nabla_\theta\log p_\theta(x_{t-1}|x_t, c) \cdot b(x_t, c, t)\right]}_{\;=:\;B\;-\;B\;}
\end{align}

Combining the first and the last term yields the advantage-weighted form,
\begin{align}
\nabla_\theta J(\theta) = \mathbb{E}_{c,\tau}\left[\sum_{t=1}^{T} \nabla_\theta\log p_\theta(x_{t-1}|x_t, c) \cdot (r(x_0, c)-b(x_t, c, t))\right] + B,
\label{eq:grad-with-B}
\end{align}
where, by definition,
\begin{align}
\nonumber B := \mathbb{E}_{c,\tau}\left[\sum_{t=1}^{T} \nabla_\theta\log p_\theta(x_{t-1}|x_t, c) \cdot b(x_t, c, t)\right]
\end{align}
Thus we have rewritten the gradient as an advantage-weighted expectation plus the extra term $B$.
To complete the proof we show $B=0$.

Write $B$ as an explicit integral over all variables.  Use the factorization
$p_\theta(\tau|c)=p(x_T)\prod_{s=1}^T p_\theta(x_{s-1}|x_s, c)$ and apply Fubini/Tonelli to
reorder integrals (assume the usual integrability and smoothness conditions).  Then
\begin{align*}
\nonumber B = \int p(c) \int \left[ \sum_{t=1}^{T} b(x_t, c, t) \nabla_\theta\log p_\theta(x_{t-1}|x_t, c)\right] 
\nonumber p_\theta(\tau|c) d\tau dc.
\end{align*}
Fix a particular time index $t$, and partition the inner integral by conditioning on $(x_t, c)$.
Denote the remaining coordinates of the trajectory (all variables except $x_{t-1}$ and $x_t$)
by ``rest''.  Then the integral over the full trajectory can be reordered as
\begin{align*}
\int  b(x_t, c, t) \nabla_\theta\log p_\theta(x_{t-1}|x_t, c) p_\theta(\tau|c) d\tau
=\int \left[ b(x_t, c, t) \left(\int p_\theta(x_{t-1}|x_t, c) 
\nabla_\theta\log p_\theta(x_{t-1}|x_t, c) dx_{t-1}\right)\right]
\\p_\theta(\text{rest}|x_t, c) d(\text{rest}) dx_t.
\end{align*}
Focus on the inner integral over $x_{t-1}$:
\begin{align*}
\int p_\theta(x_{t-1}|x_t, c)
\nabla_\theta\log p_\theta(x_{t-1}|x_t, c) dx_{t-1}
=
\int \nabla_\theta p_\theta(x_{t-1}|x_t, c) dx_{t-1}
&=
\\[4pt]
\nabla_\theta \int p_\theta(x_{t-1}|x_t, c) dx_{t-1}
=
\nabla_\theta 1
&= 0.
\end{align*}
Because this inner integral equals zero for every fixed $(x_t, c)$, the entire integrand above is zero,
so the contribution of index $t$ to $B$ vanishes.  Summing over $t=1,\dots,T$ therefore yields
\[
B = 0
\]

\subsection{Advantage-weighted gradient form and Monte Carlo estimator}
Since $B=0$, equation \eqref{eq:grad-with-B} reduces to the unbiased advantage-weighted gradient
identity.
\begin{align*}
\nabla_\theta J(\theta)
=
\mathbb{E}_{c \sim p(c)}
\mathbb{E}_{\tau \sim p_\theta(\cdot|c)}\left[\sum_{t=1}^{T}
\nabla_\theta\log p_\theta(x_{t-1}|x_t, c) 
(r(x_0, c)-b(x_t, c, t))\right]
\end{align*}
A standard, unbiased Monte Carlo estimator based on $N$ sampled trajectories
$\{\tau^{(n)}, c^{(n)}\}_{n=1}^N$ is
\begin{align*}
\widehat{\nabla_\theta J} = \frac{1}{N}\sum_{n=1}^N \sum_{t=1}^{T} \left[\nabla_\theta\log p_\theta(x_{t-1}^{(n)}|x_t^{(n)}, c^{(n)})
(r(x_0^{(n)}, c^{(n)}) - b(x_t^{(n)}, c^{(n)}, t))\right]
\end{align*}

\subsection{Practical choice of a baseline function.}
The variance of the estimator at each step $t$ is minimized (for a scalar baseline) by the optimal choice
\[
b_t^{\star}(x_t, c) = \mathbb{E}\left[r(x_0, c) | x_t, c, t\right],
\]
the conditional expectation of the return given $(x_t, c, t)$ (standard result by orthogonal projection in $L^2$; derive by setting $\partial\,\mathrm{Var}/\partial b_t=0$).

Thus choosing $b(x_t, c, t)=V_\phi(x_t, c, t)$, precisely the value function in \eqref{eq:value-def}, and training $V_\phi$ to regress the observed terminal reward
$r(x_0, c)$ (Monte Carlo regression) approximates the optimal least-squares baseline $b^\star(x_t)=\mathbb{E}[r(x_0, c)|x_t]$ and thereby reduces variance of the estimator without introducing bias.

\subsection{Off-policy extension via importance sampling}
To enable multiple optimization steps on reused trajectories, we consider data collected under an earlier policy $p_{\theta_{\text{old}}}$. Correcting distribution shift with per-step importance weights yields the off-policy advantage-weighted estimator:
\begin{align*}
\nabla_\theta J(\theta) = \mathbb{E}_{c,\tau \sim p_{\theta_{\text{old}}}}\left[\sum_{t=1}^{T} \frac{p_\theta(x_{t-1}|x_t, c)}{p_{\theta_{\text{old}}}(x_{t-1}|x_t, c)} 
\nabla_\theta\log p_\theta(x_{t-1}|x_t, c)\,(r(x_0, c)-b(x_t, c, t))\right].
\end{align*}
This follows by standard importance sampling arguments \cite{10.5555/645531.656005} applied to each term of the score-function expansion.

\section{Additional Experimental Results and Evaluation Protocol Details}
\label{app:additional_results}

\subsection{UnlearnCanvas Evaluation Protocol Details}
\label{subsec:eval_protocol}
For each target concept, we evaluate the corresponding unlearned model on a fixed prompt grid combining \emph{content} and \emph{style}. Concretely, we use the template \textit{``A \{class\_name\} image in \{style\} style''}. For each of the 20 object classes and each of the 51 available styles (e.g., \textit{Watercolor}, \textit{Cubism}, \textit{Van\_Gogh}, \textit{Sketch}, \textit{Neon\_Lines}), we generate 20 images, yielding \(20 \times 51 \times 20\) samples per unlearned model. We run the UnlearnCanvas ViT classifier on each image and aggregate predicted class probabilities to compute per-class UA (for the target class) and IRA (mean over all non-target classes).

\begin{table}[h]
\centering
\caption{\textbf{Per-class metrics on UnlearnCanvas for CGRU.} (higher is better).}
\label{tab:per_class_metrics}
\small
\begin{tabular}{l|cc}
\toprule
\textbf{Target class} & \textbf{UA} ($\uparrow$) & \textbf{IRA} ($\uparrow$) \\
\midrule
Architectures & 94.50\% & 78.11\% \\
Bears & 97.94\% & 72.53\% \\
Birds & 94.29\% & 85.34\% \\
Butterfly & 96.46\% & 78.91\% \\
Cats & 95.33\% & 84.30\% \\
Dogs & 98.04\% & 75.64\% \\
Fishes & 97.25\% & 64.76\% \\
Flame & 98.43\% & 83.67\% \\
Flowers & 91.63\% & 89.78\% \\
Frogs & 96.08\% & 71.51\% \\
Horses & 98.71\% & 79.16\% \\
Humans & 95.28\% & 82.19\% \\
Jellyfish & 90.19\% & 81.19\% \\
Rabbits & 93.87\% & 96.42\% \\
Sandwiches & 92.58\% & 88.12\% \\
Sea & 95.98\% & 81.98\% \\
Statues & 96.22\% & 85.77\% \\
Towers & 94.27\% & 83.26\% \\
Trees & 93.51\% & 81.51\% \\
Waterfalls & 96.86\% & 88.75\% \\
\bottomrule
\end{tabular}
\end{table}

\subsection{Misclassification matrices.}
\begin{figure}[h]
\centering
\includegraphics[width=0.24\textwidth]{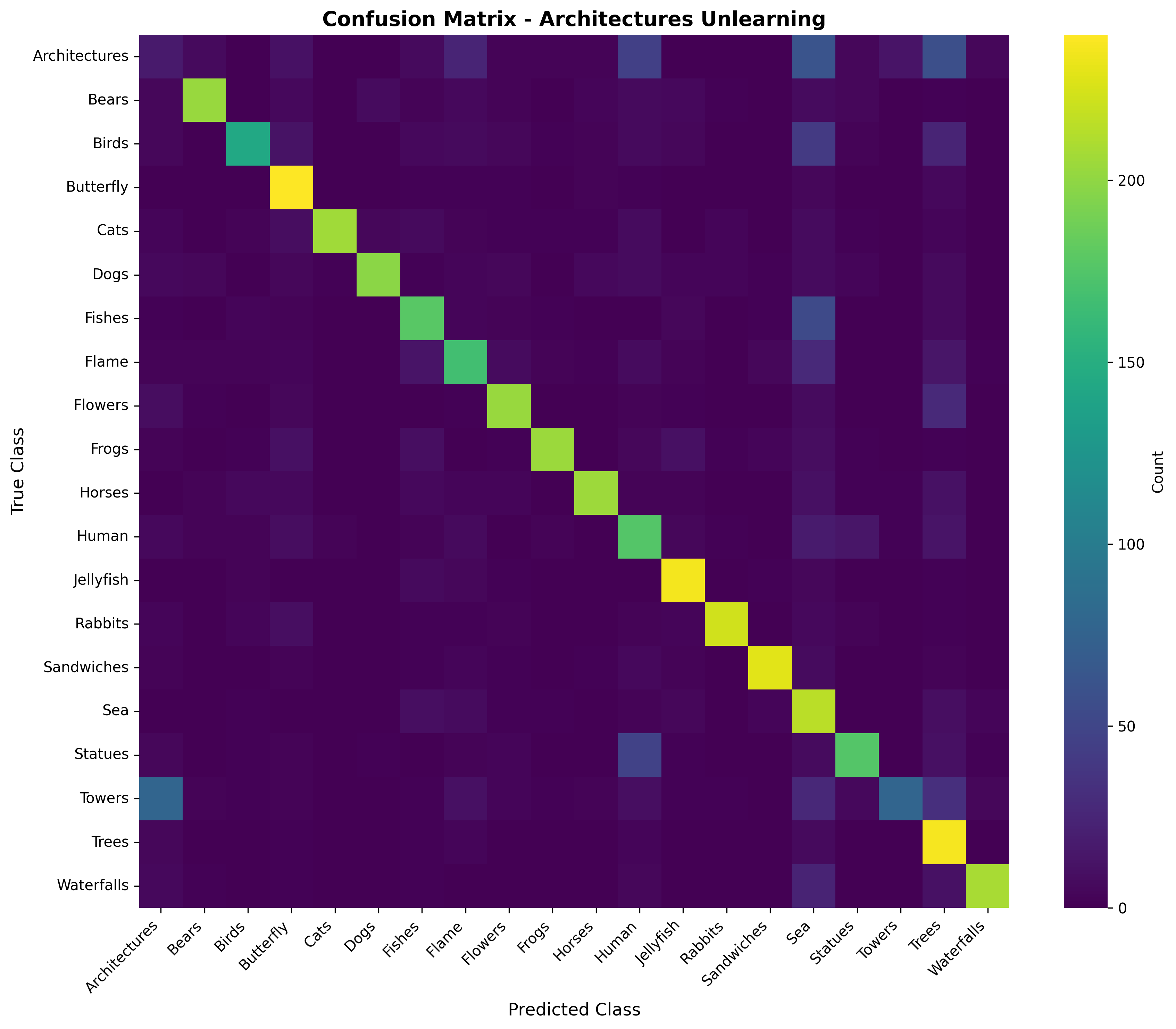}\hfill
\includegraphics[width=0.24\textwidth]{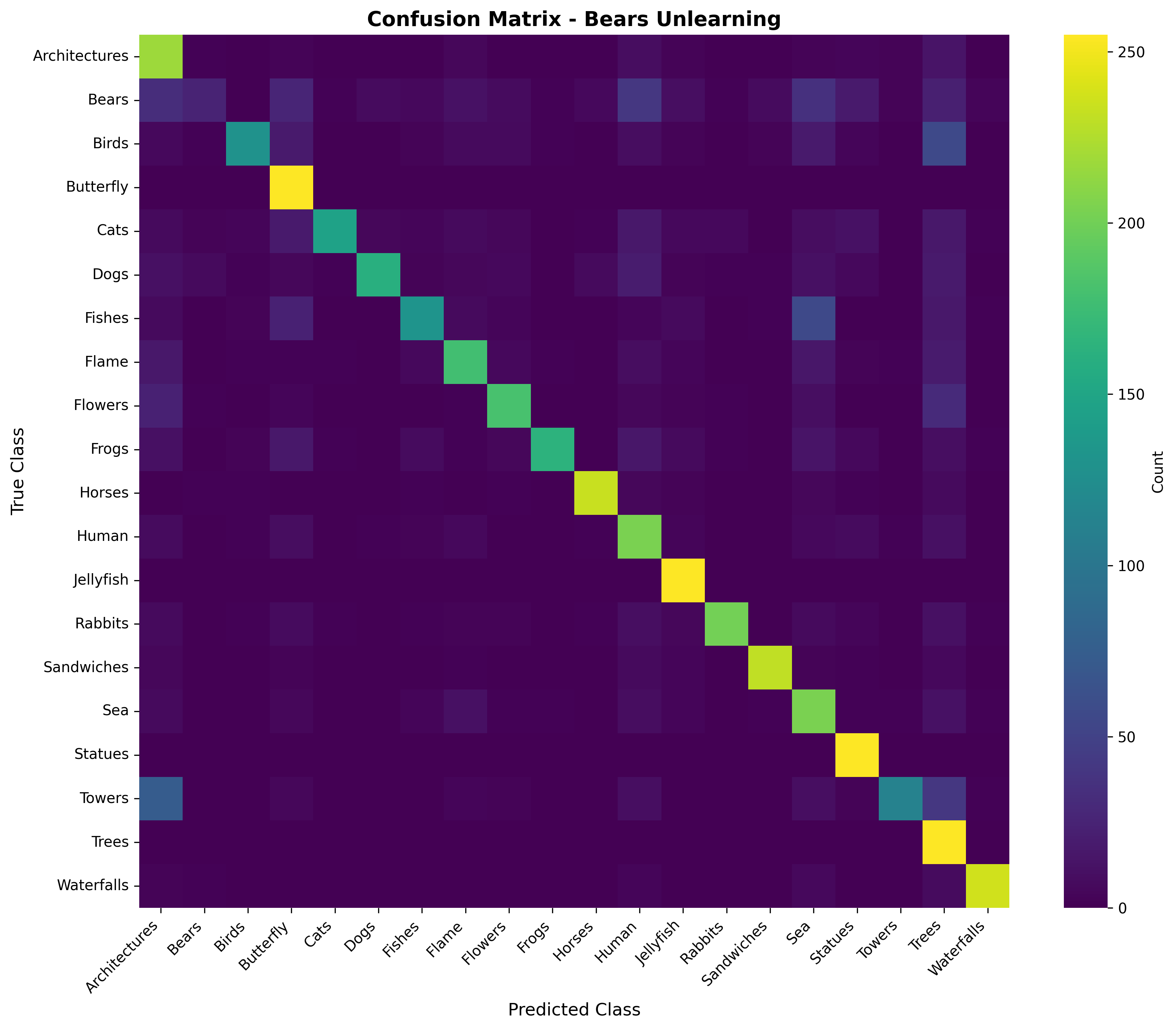}\hfill
\includegraphics[width=0.24\textwidth]{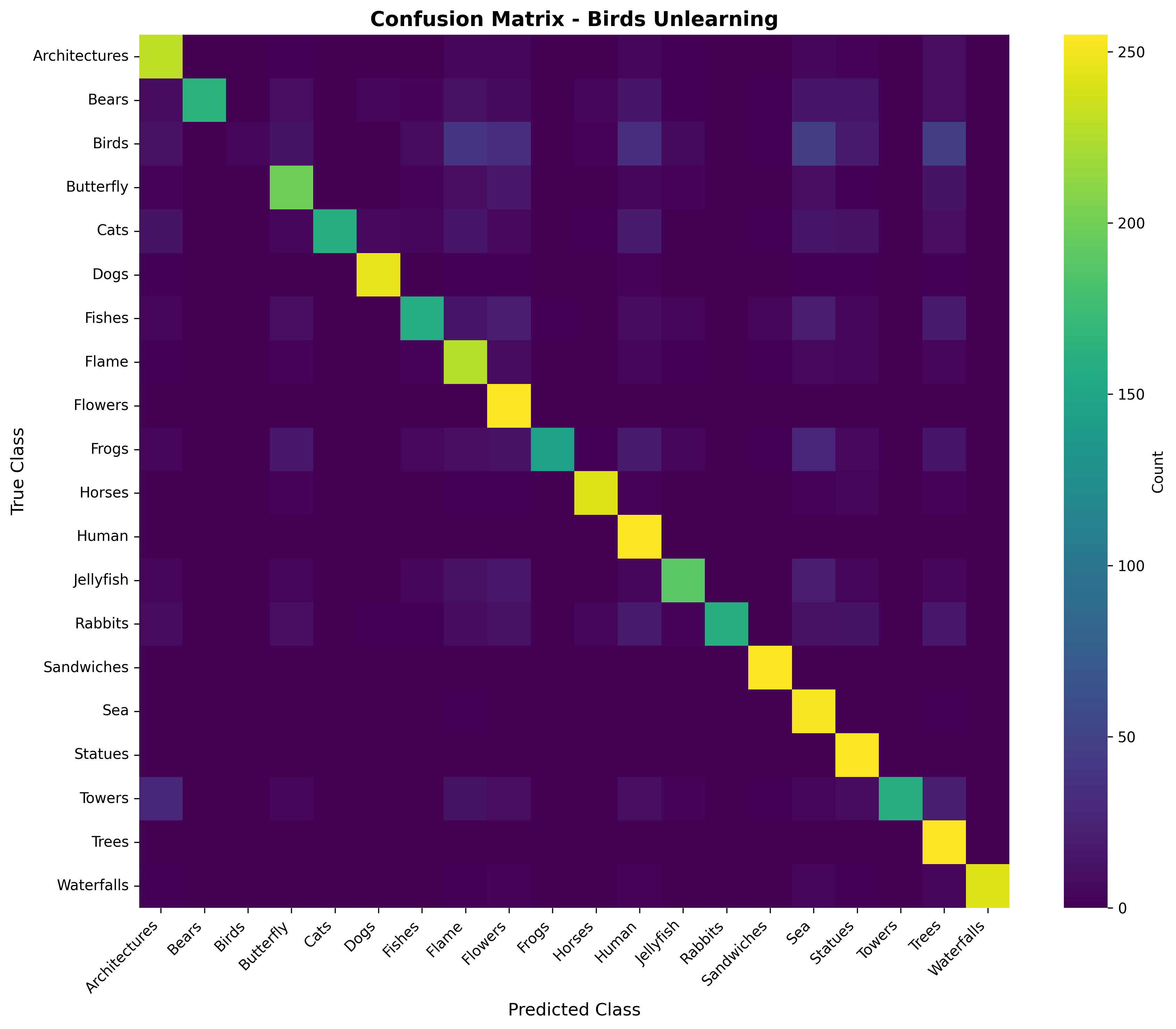}\hfill
\includegraphics[width=0.24\textwidth]{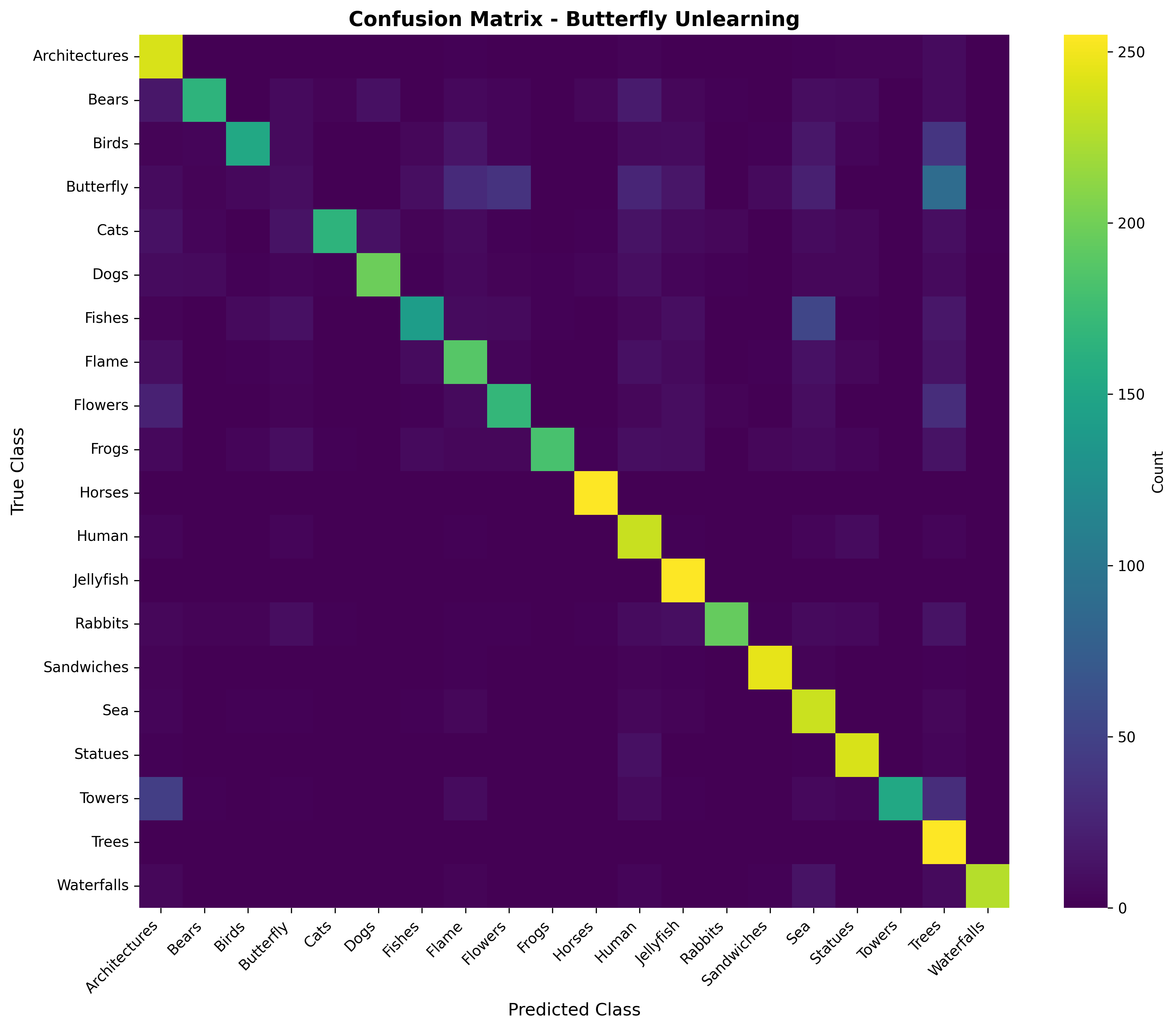}\\[0.6em]
\includegraphics[width=0.24\textwidth]{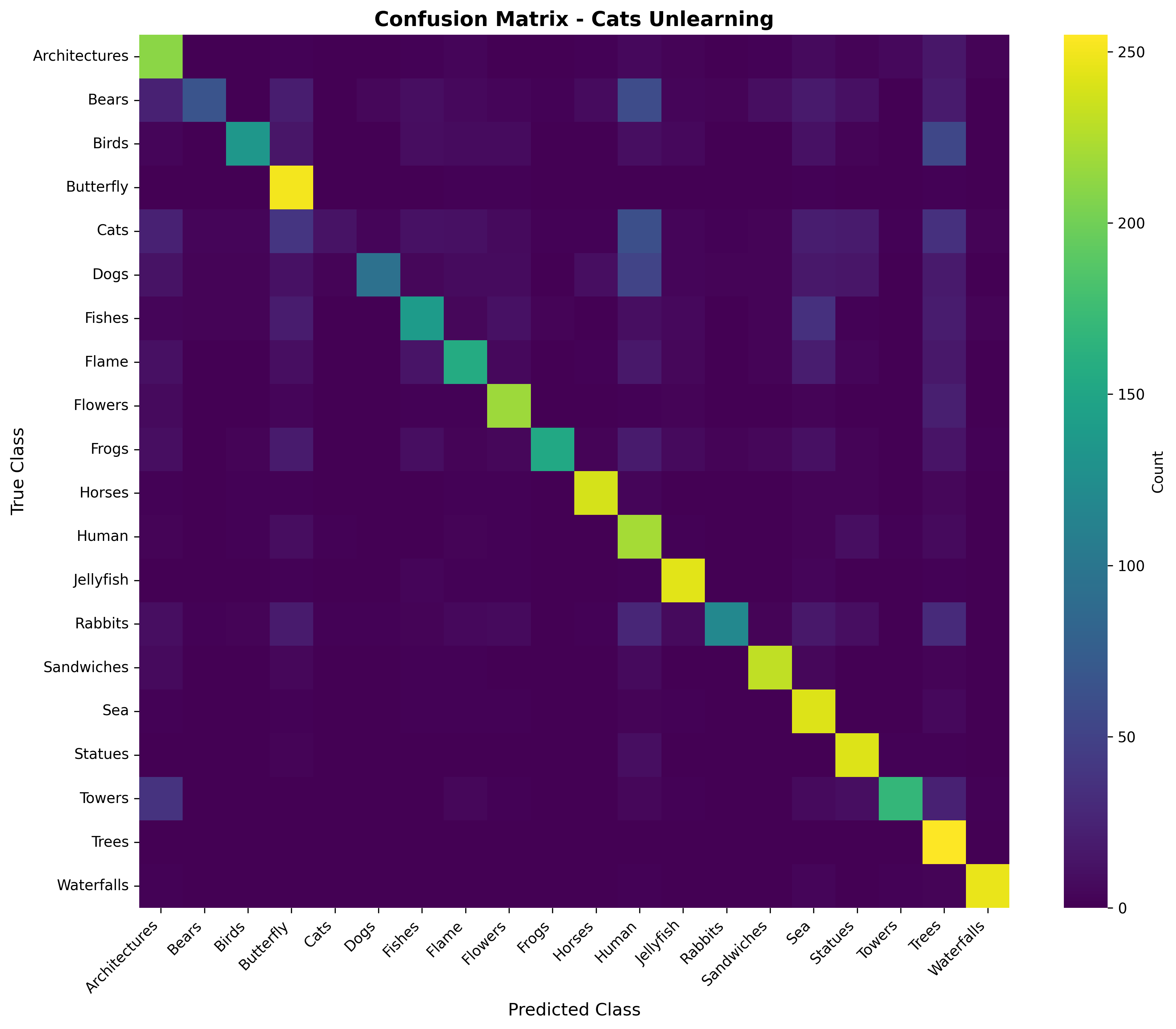}\hfill
\includegraphics[width=0.24\textwidth]{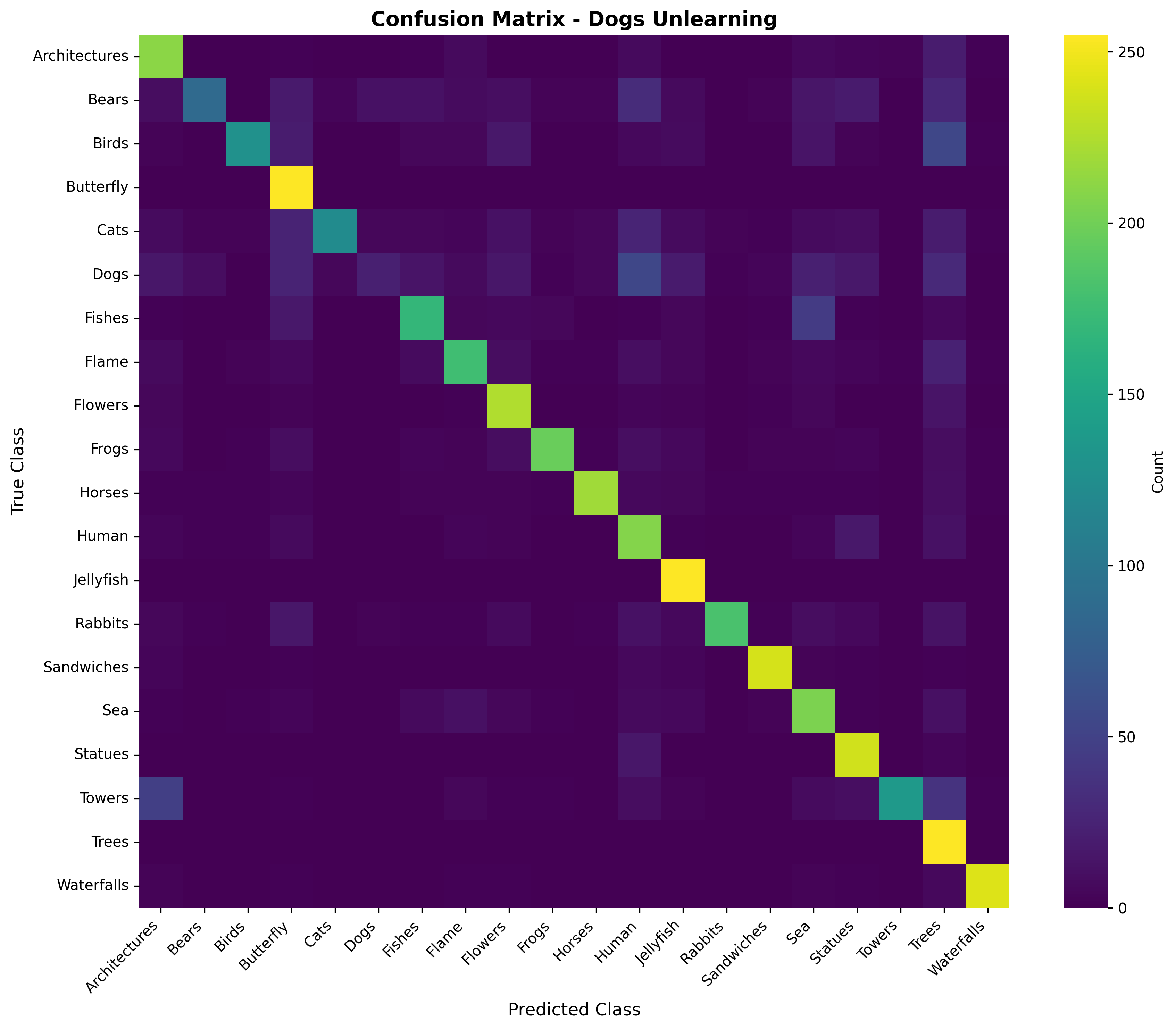}\hfill
\includegraphics[width=0.24\textwidth]{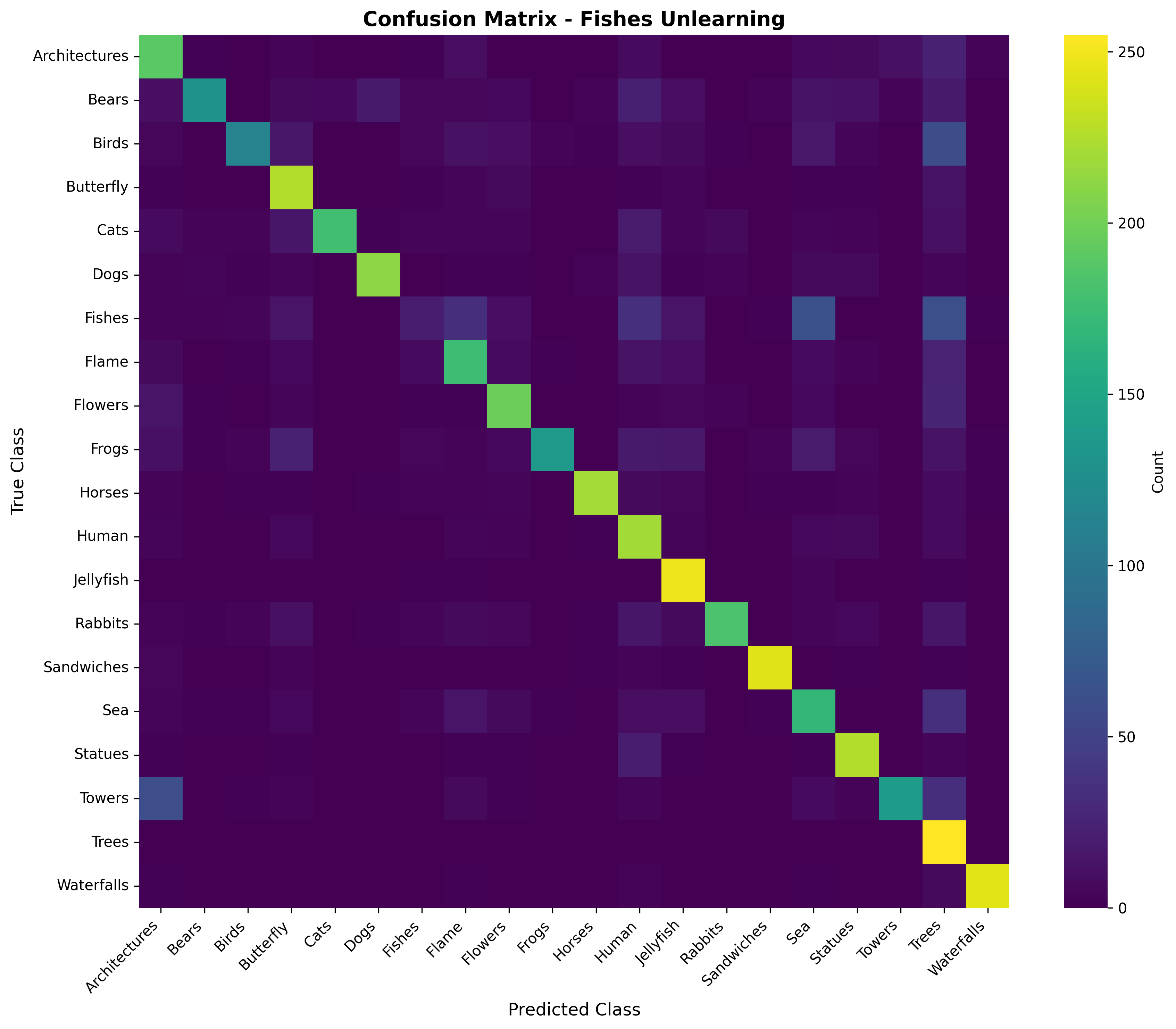}\hfill
\includegraphics[width=0.24\textwidth]{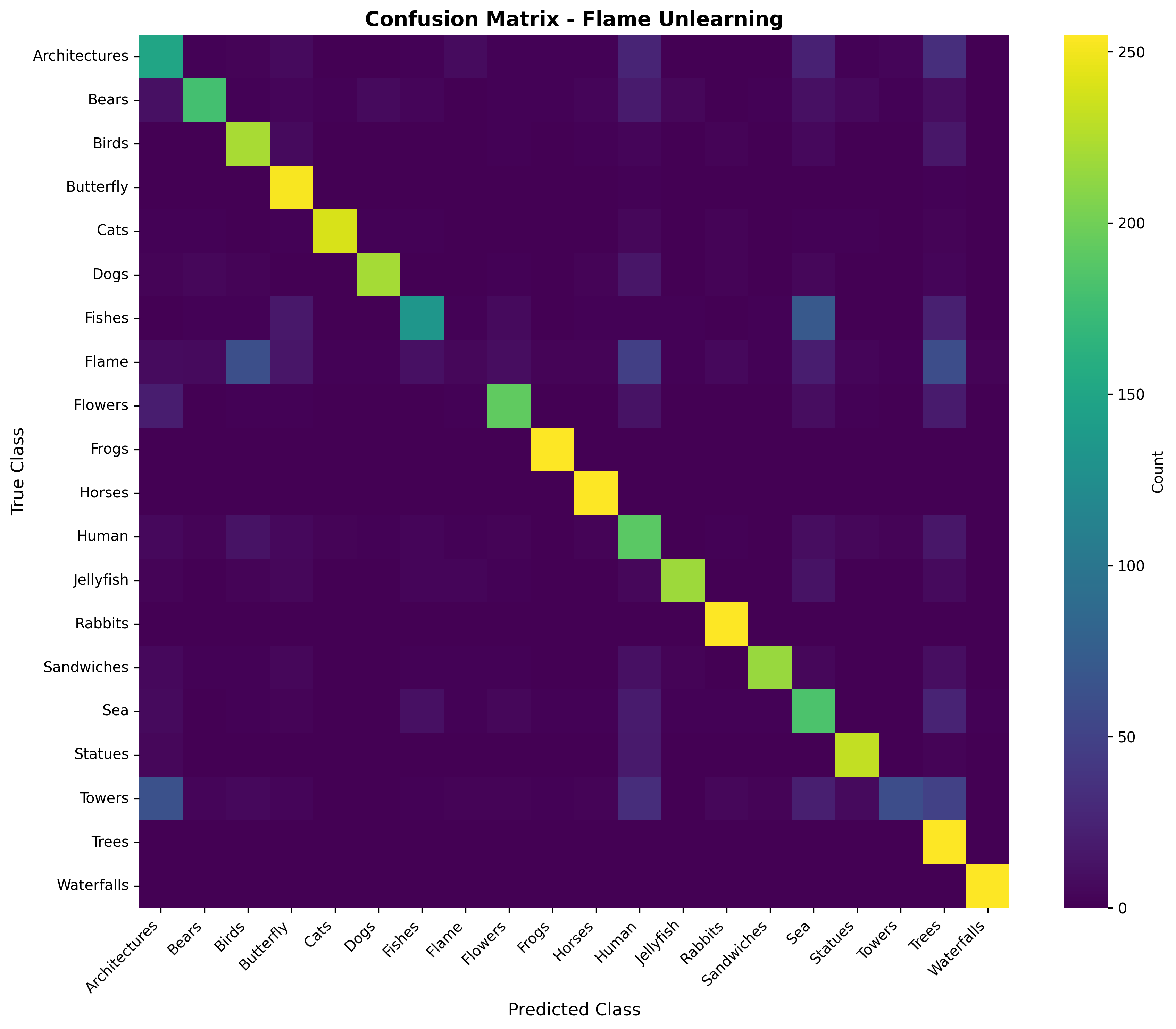}\\[0.6em]
\caption{\textbf{Per-class misclassification matrices (1/2).} One matrix per target class.}
\label{fig:misclassification_matrices}
\end{figure}

\begin{figure}[t]
\centering
\includegraphics[width=0.33\textwidth]{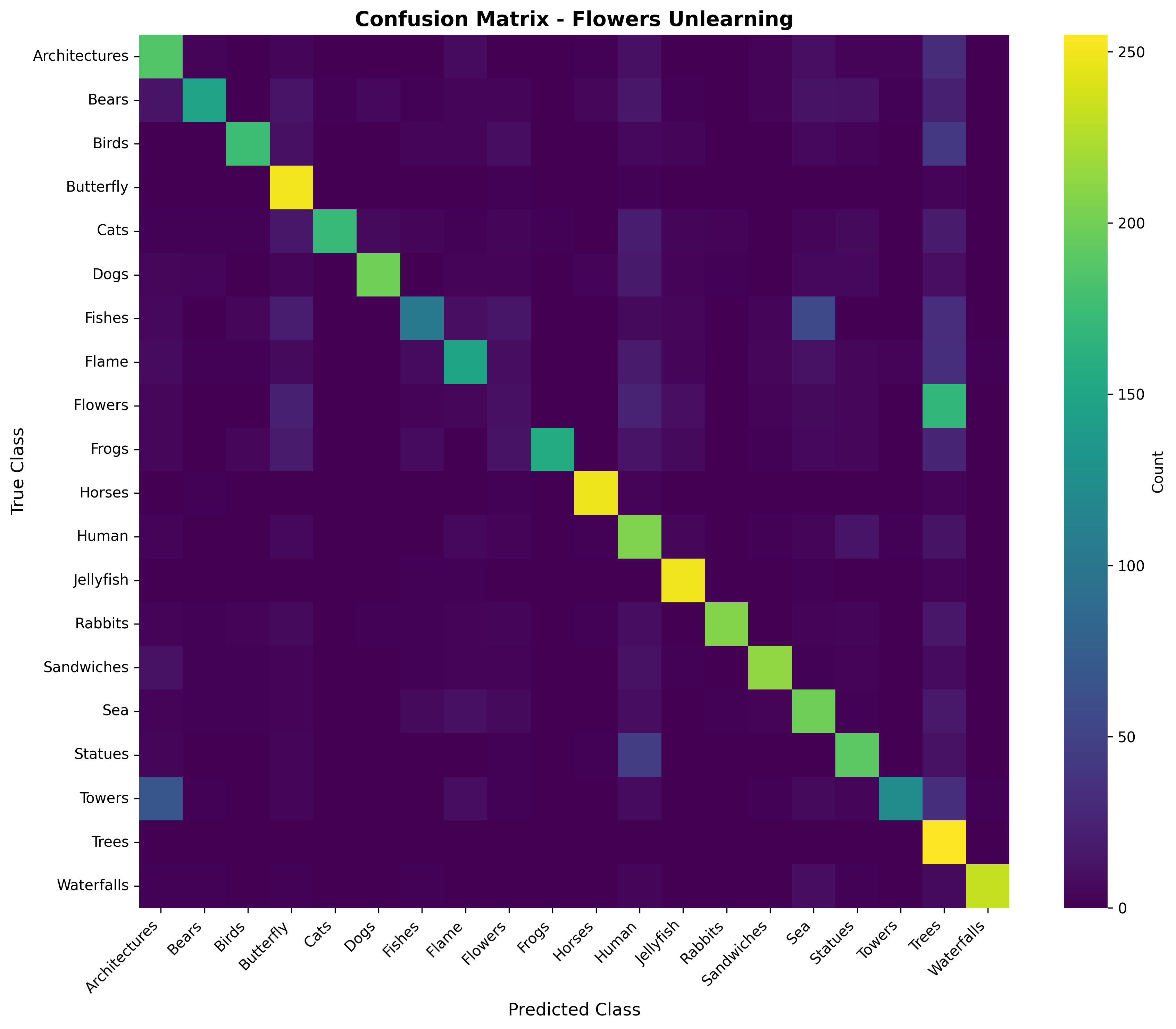}\hfill
\includegraphics[width=0.33\textwidth]{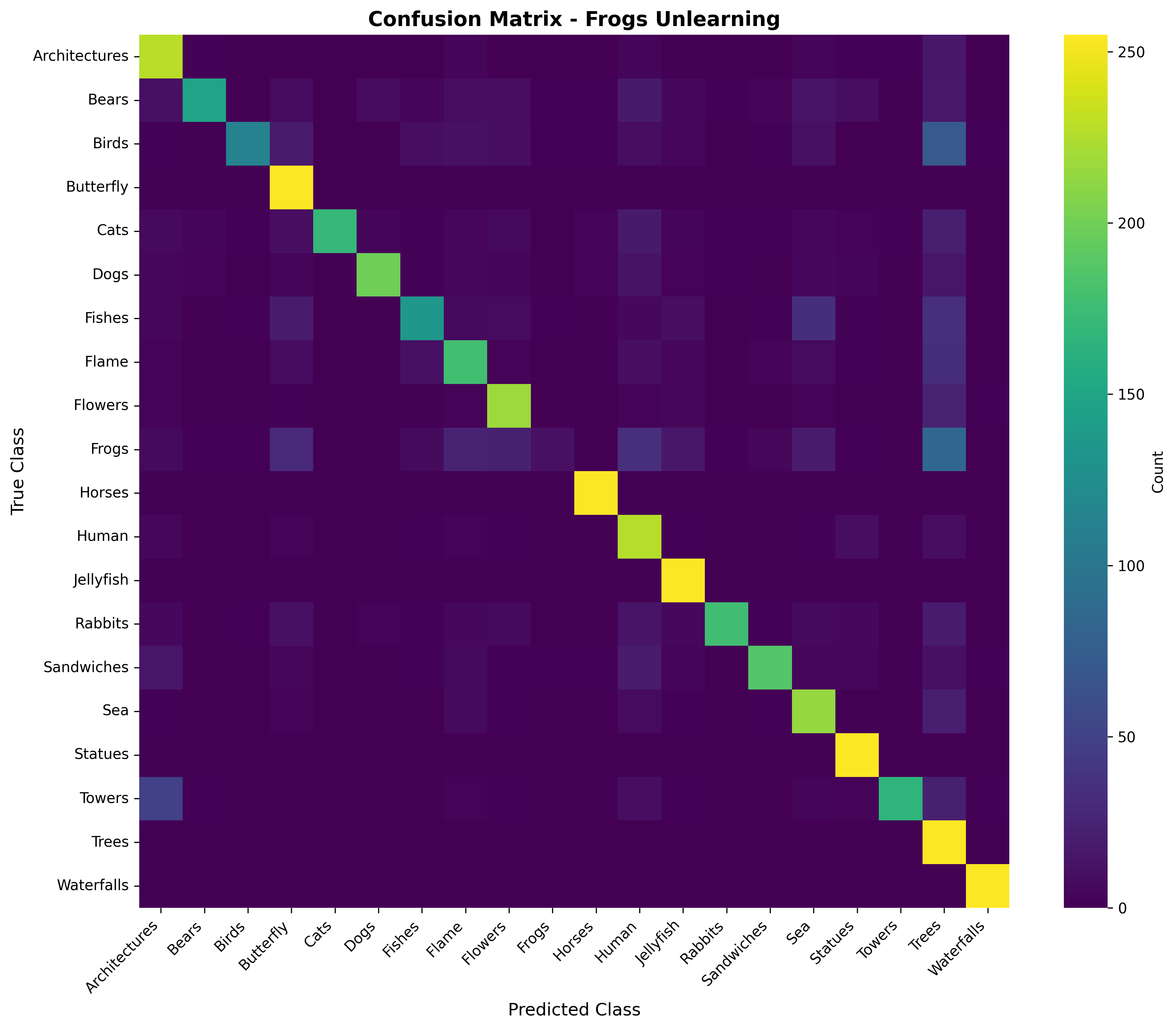}\hfill
\includegraphics[width=0.33\textwidth]{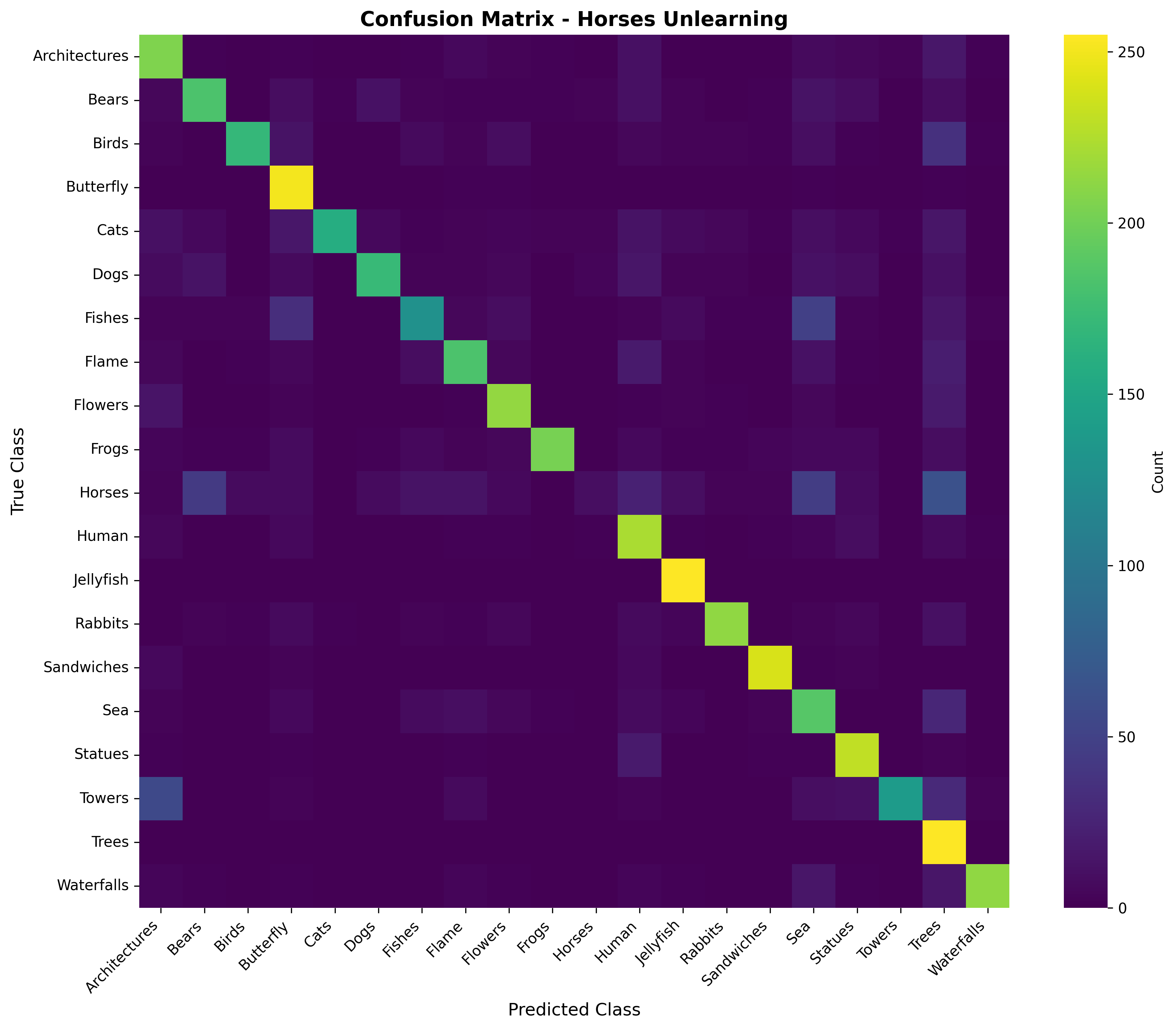}\\[0.6em]
\includegraphics[width=0.33\textwidth]{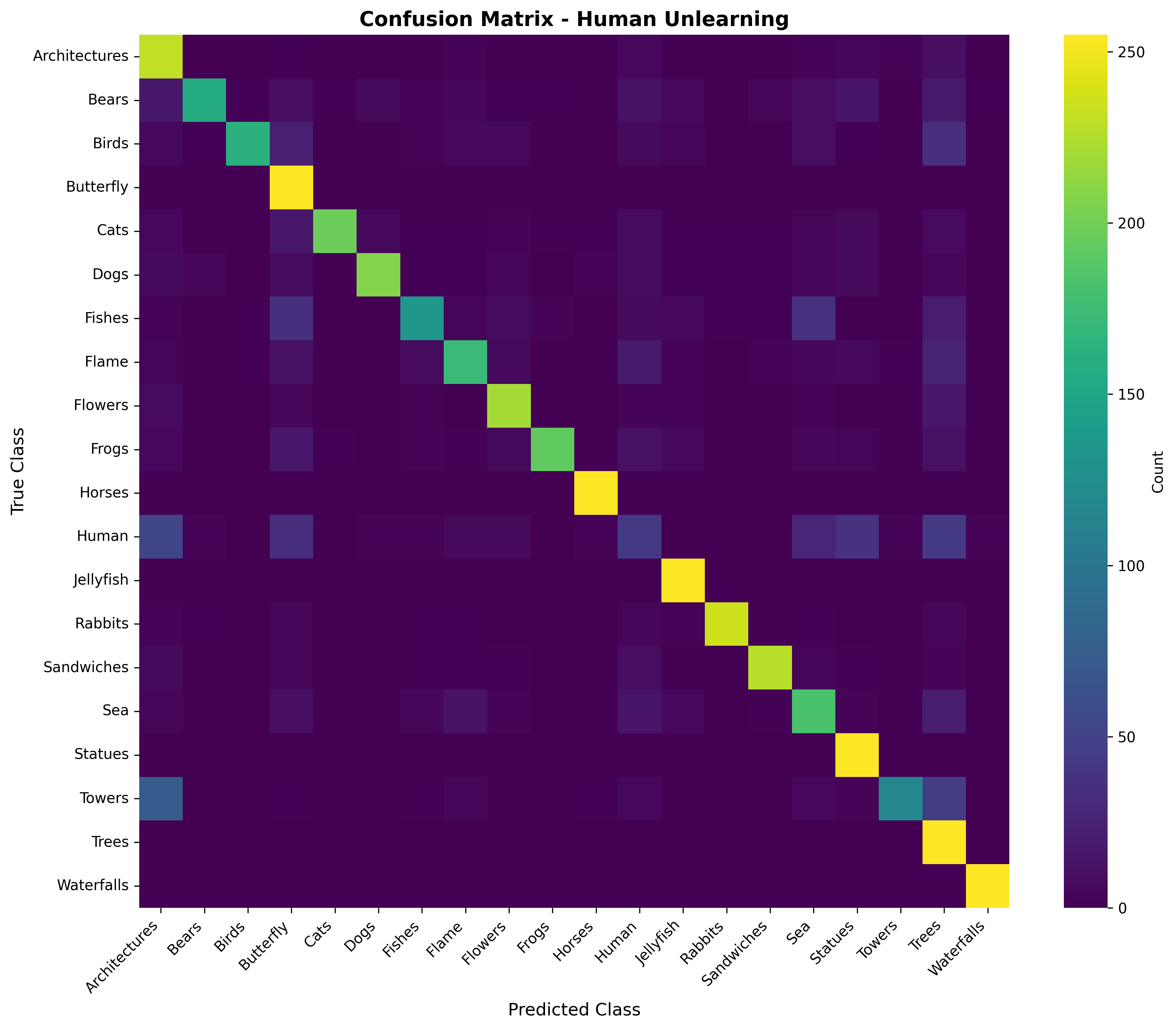}\hfill
\includegraphics[width=0.33\textwidth]{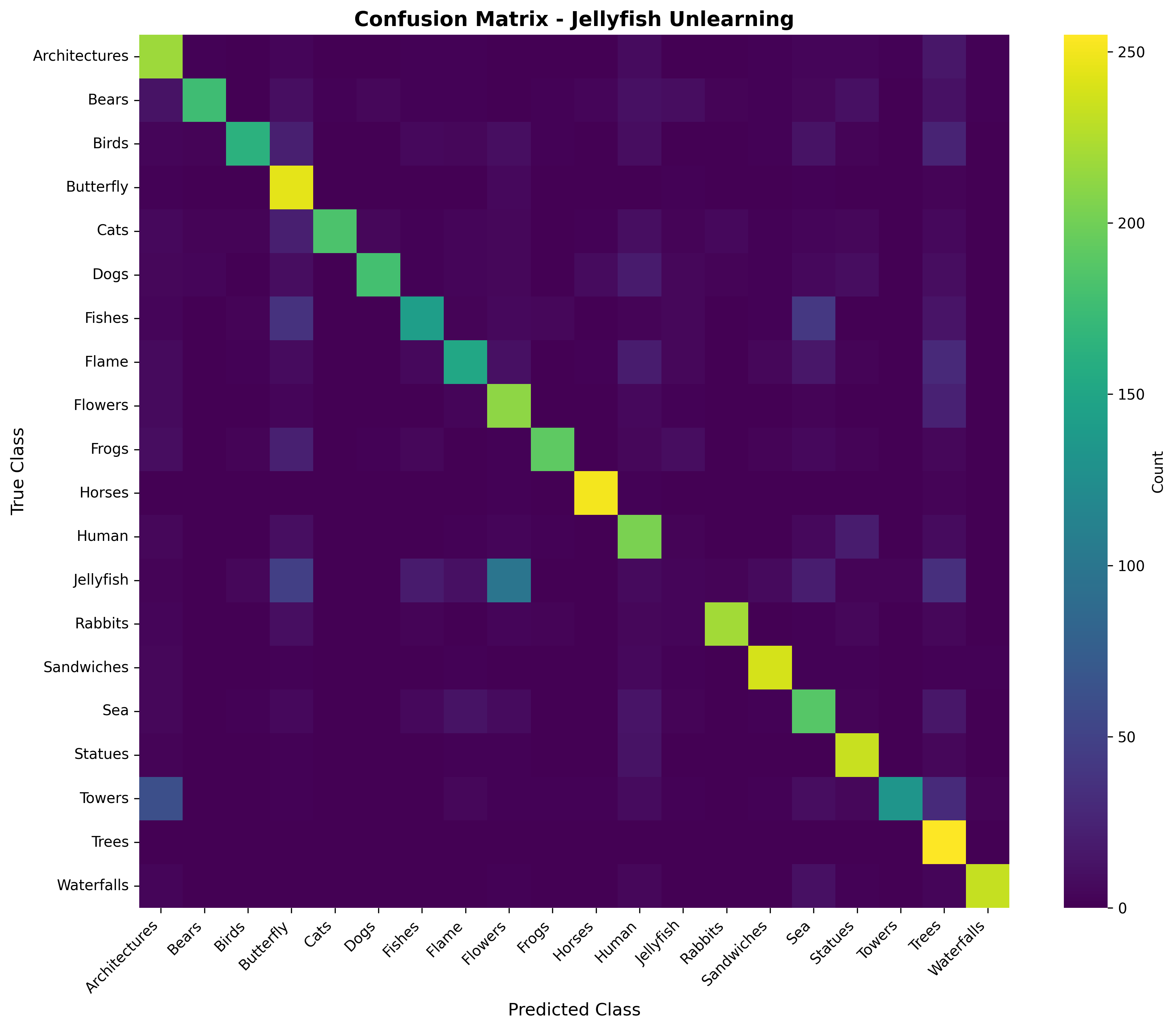}\hfill
\includegraphics[width=0.33\textwidth]{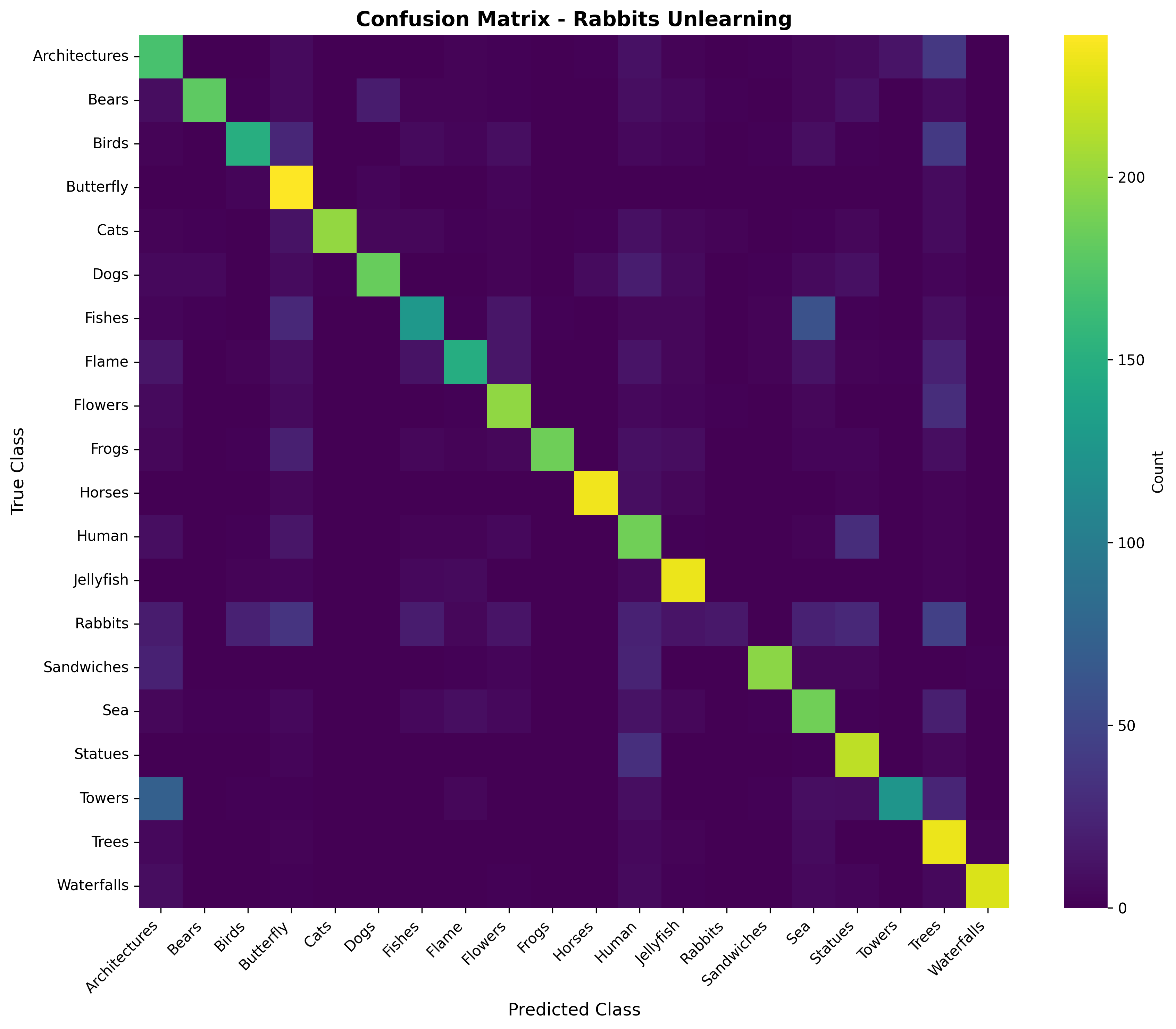}\\[0.6em]
\includegraphics[width=0.33\textwidth]{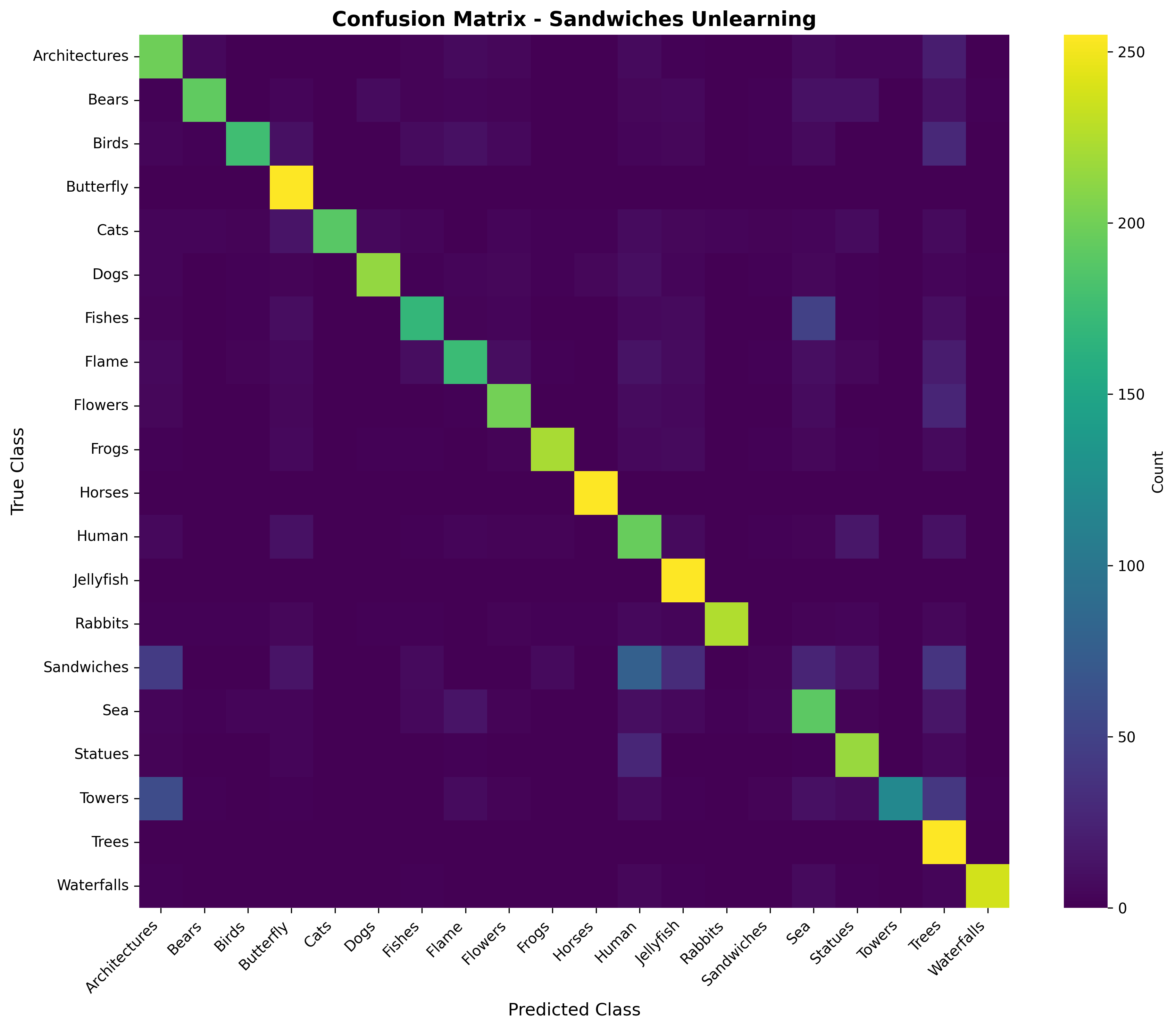}\hfill
\includegraphics[width=0.33\textwidth]{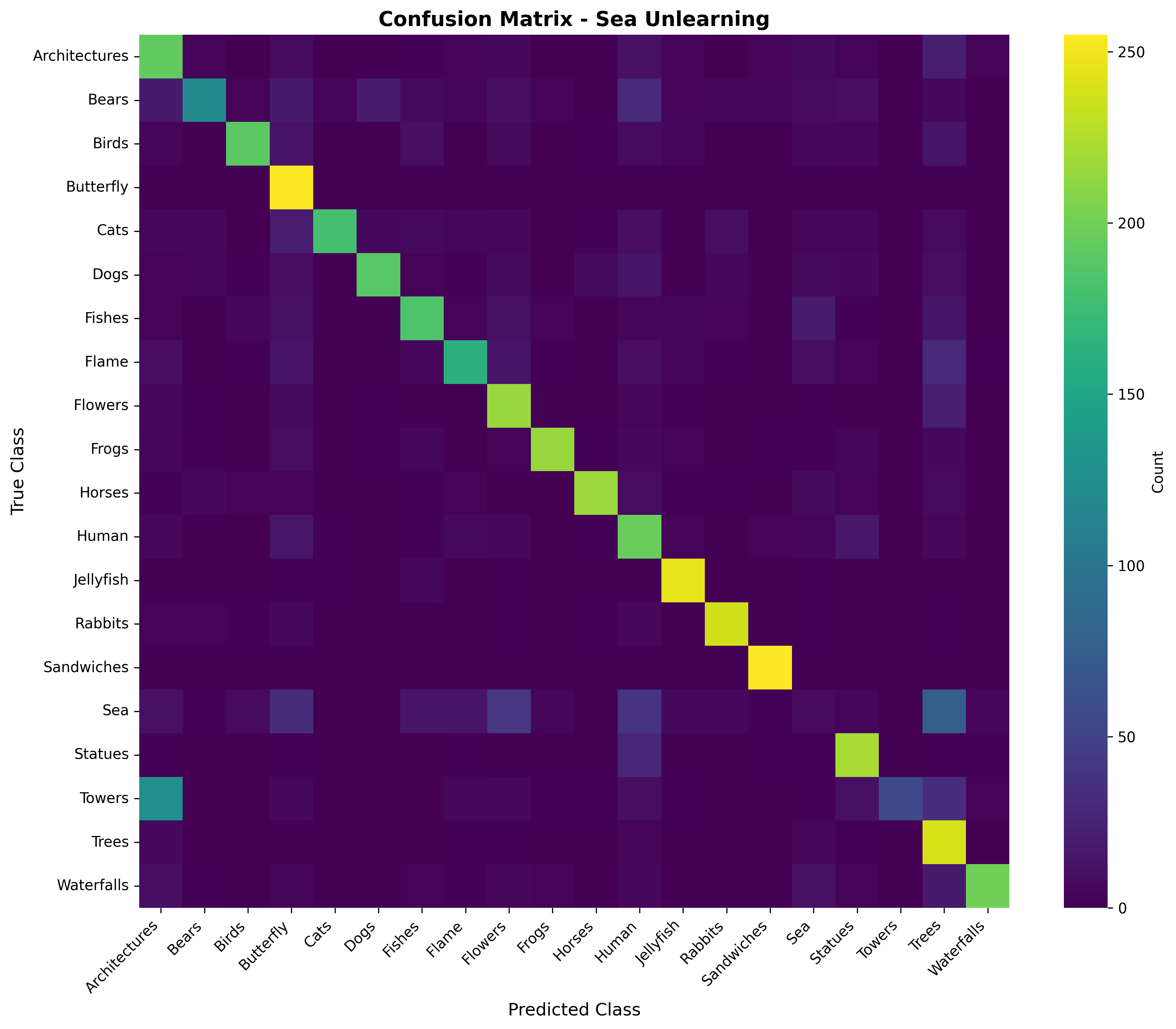}\hfill
\includegraphics[width=0.33\textwidth]{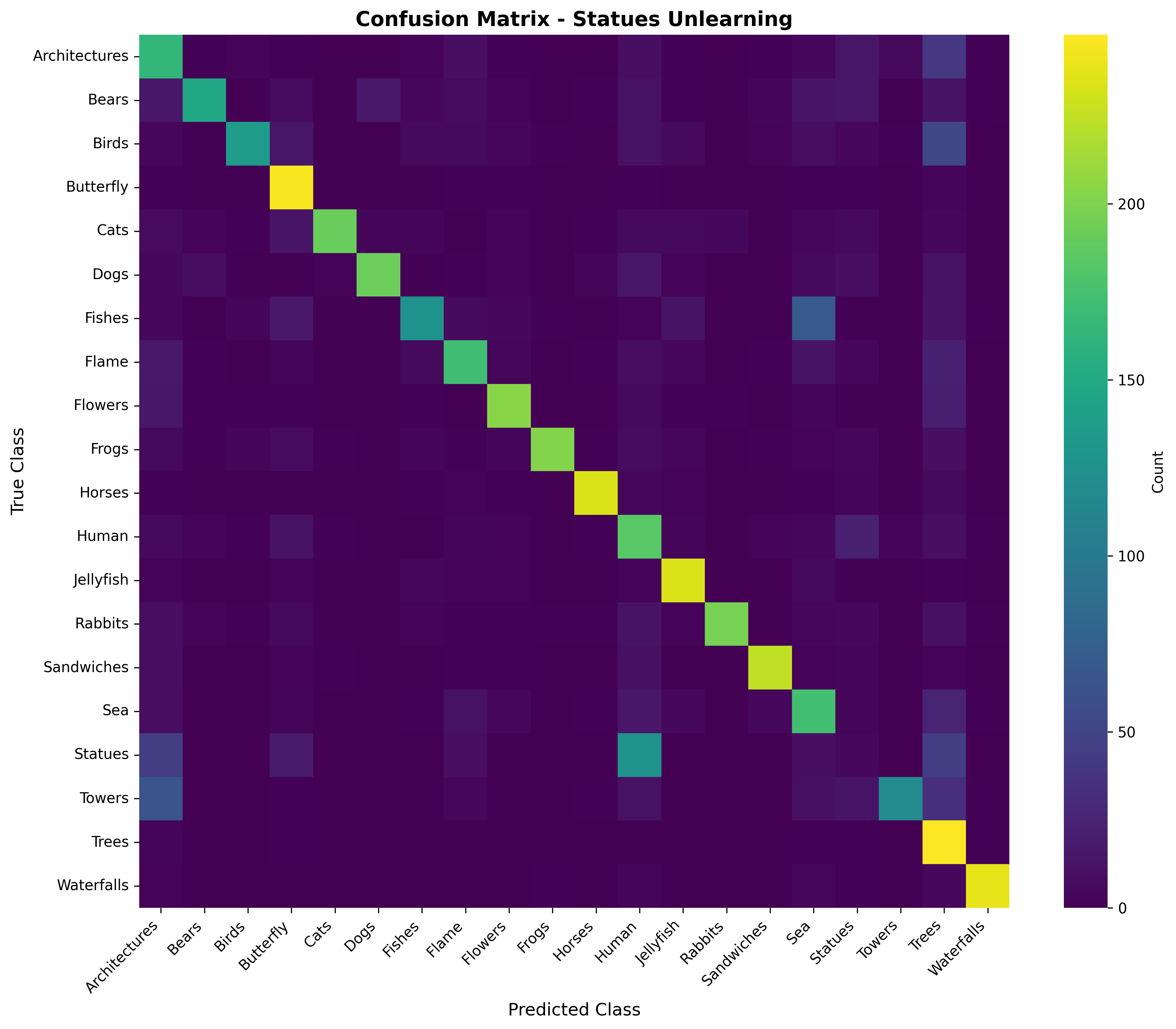}\\[0.6em]
\includegraphics[width=0.33\textwidth]{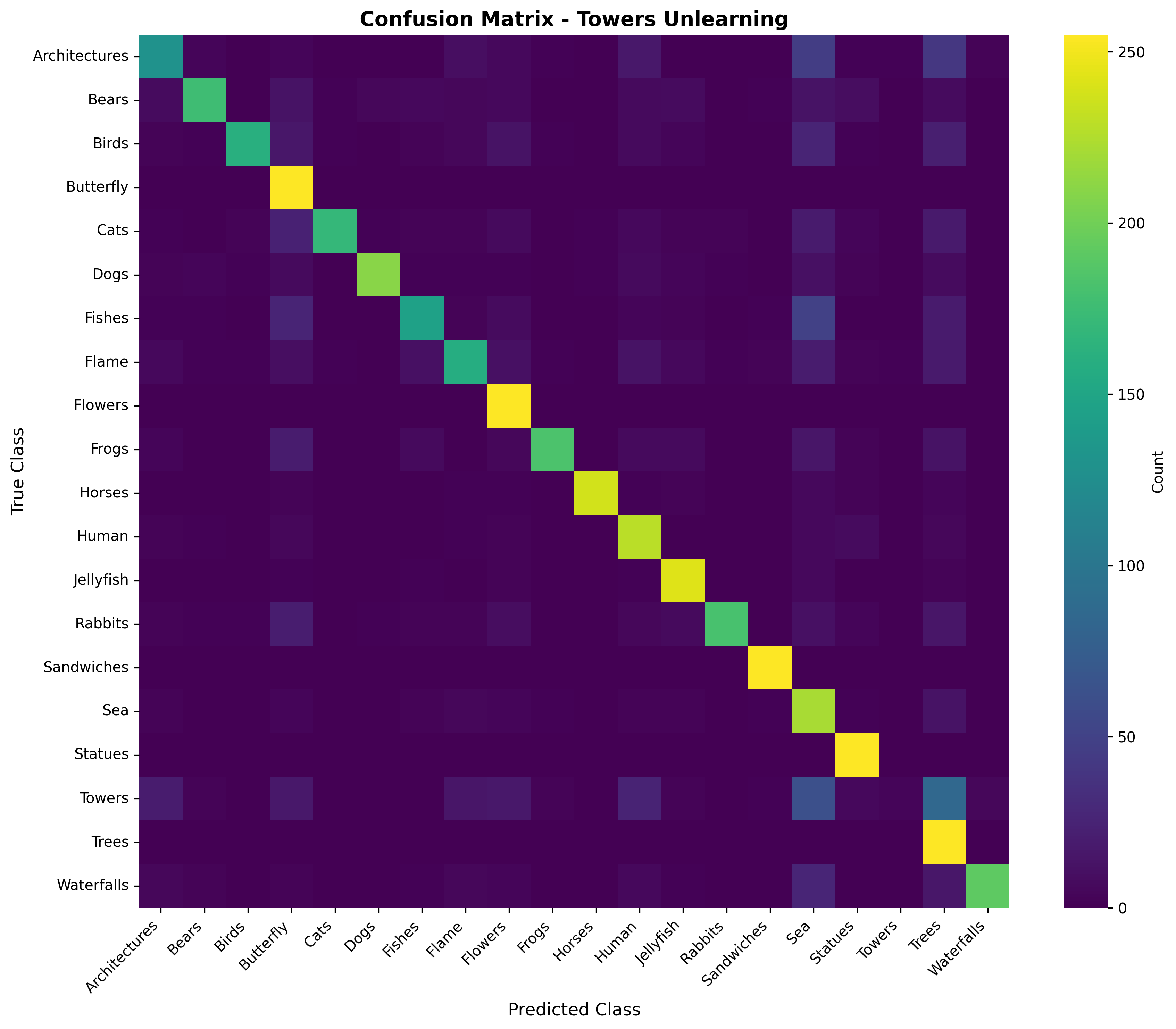}\hfill
\includegraphics[width=0.33\textwidth]{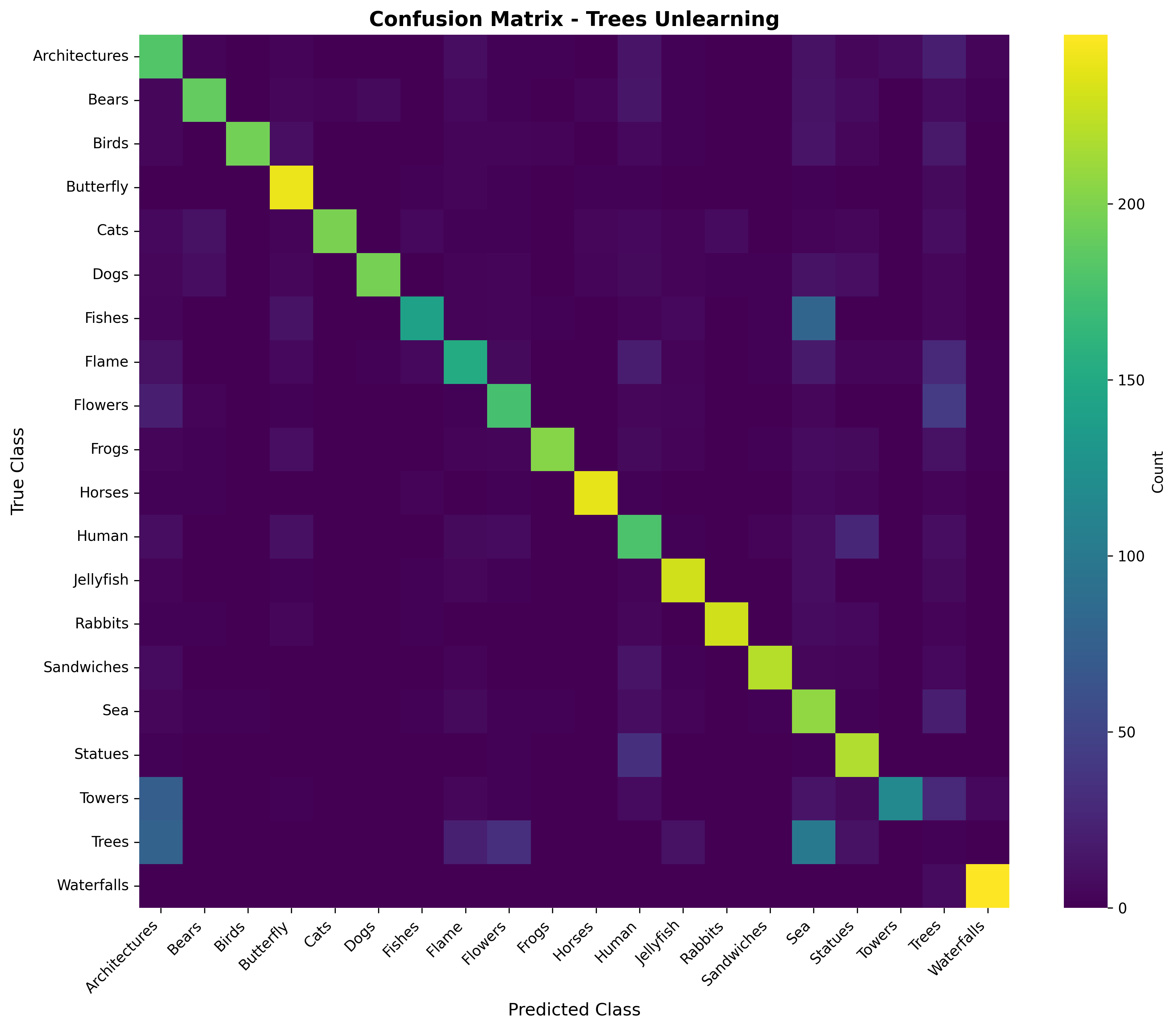}\hfill
\includegraphics[width=0.33\textwidth]{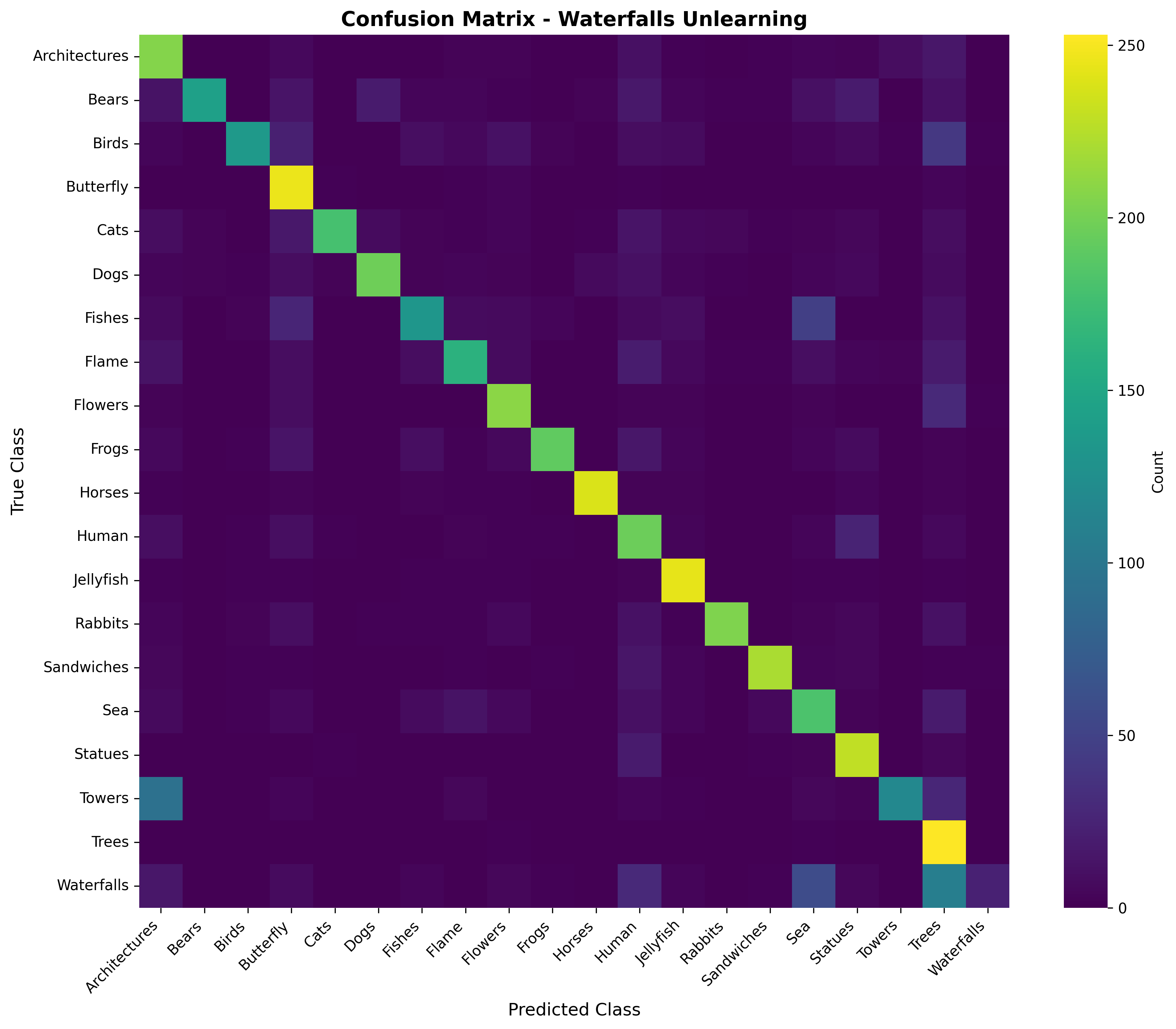}
\caption{\textbf{Per-class misclassification matrices (2/2).}}
\label{fig:misclassification_matrices_2}
\end{figure}

\clearpage

\subsection{Additional Qualitative Results}
\label{subsec:qual_progression}
We provide additional qualitative examples in the same format as Figure~\ref{fig:forgetting_progression} in the main paper.

\begin{figure}[h]
    \centering
    \includegraphics[width=\textwidth]{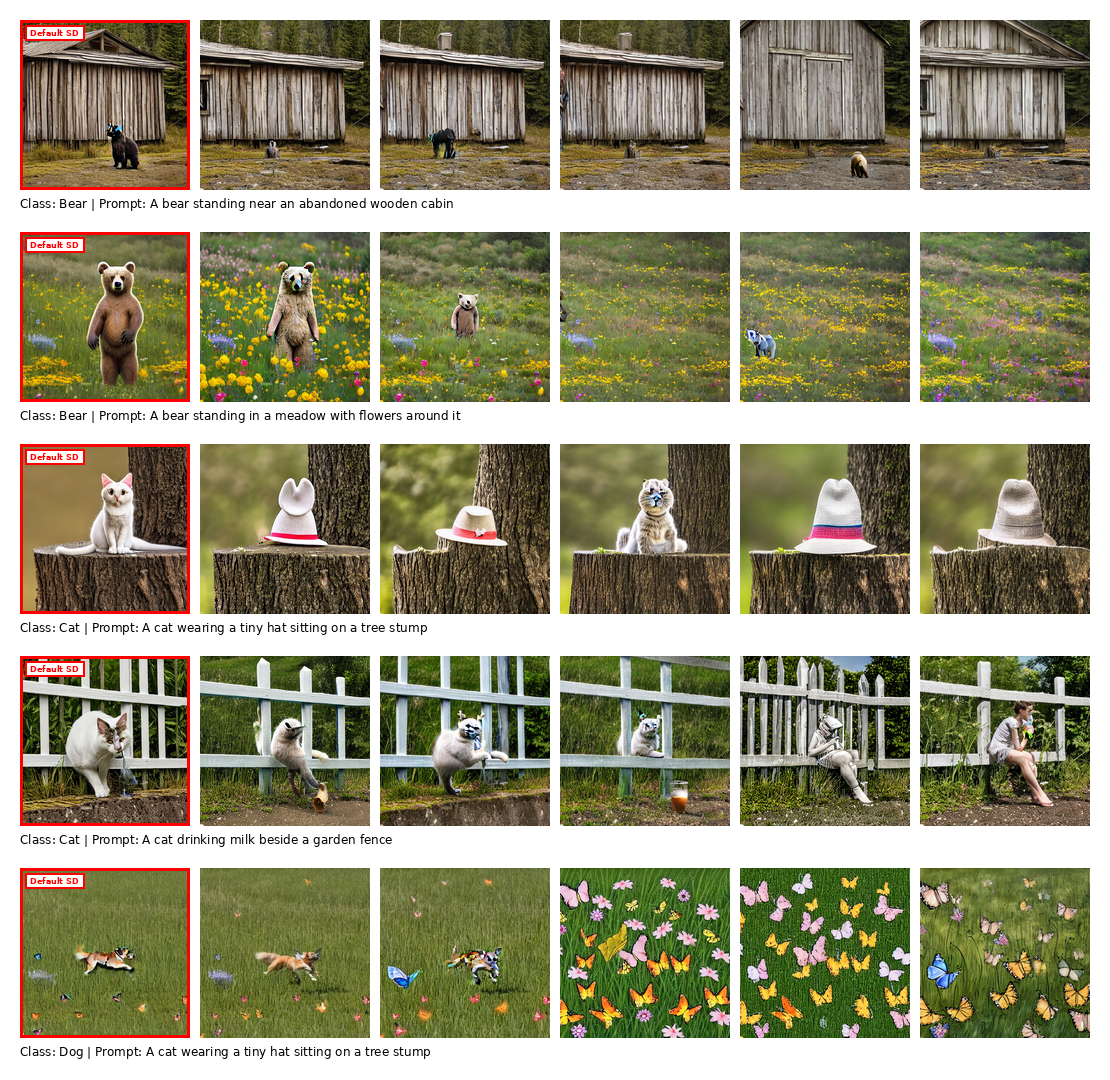}
    \caption{Qualitative progression examples (set 1).}
    \label{fig:progression_appendix_1}
\end{figure}

\begin{figure}[h]
    \centering
    \includegraphics[width=\textwidth]{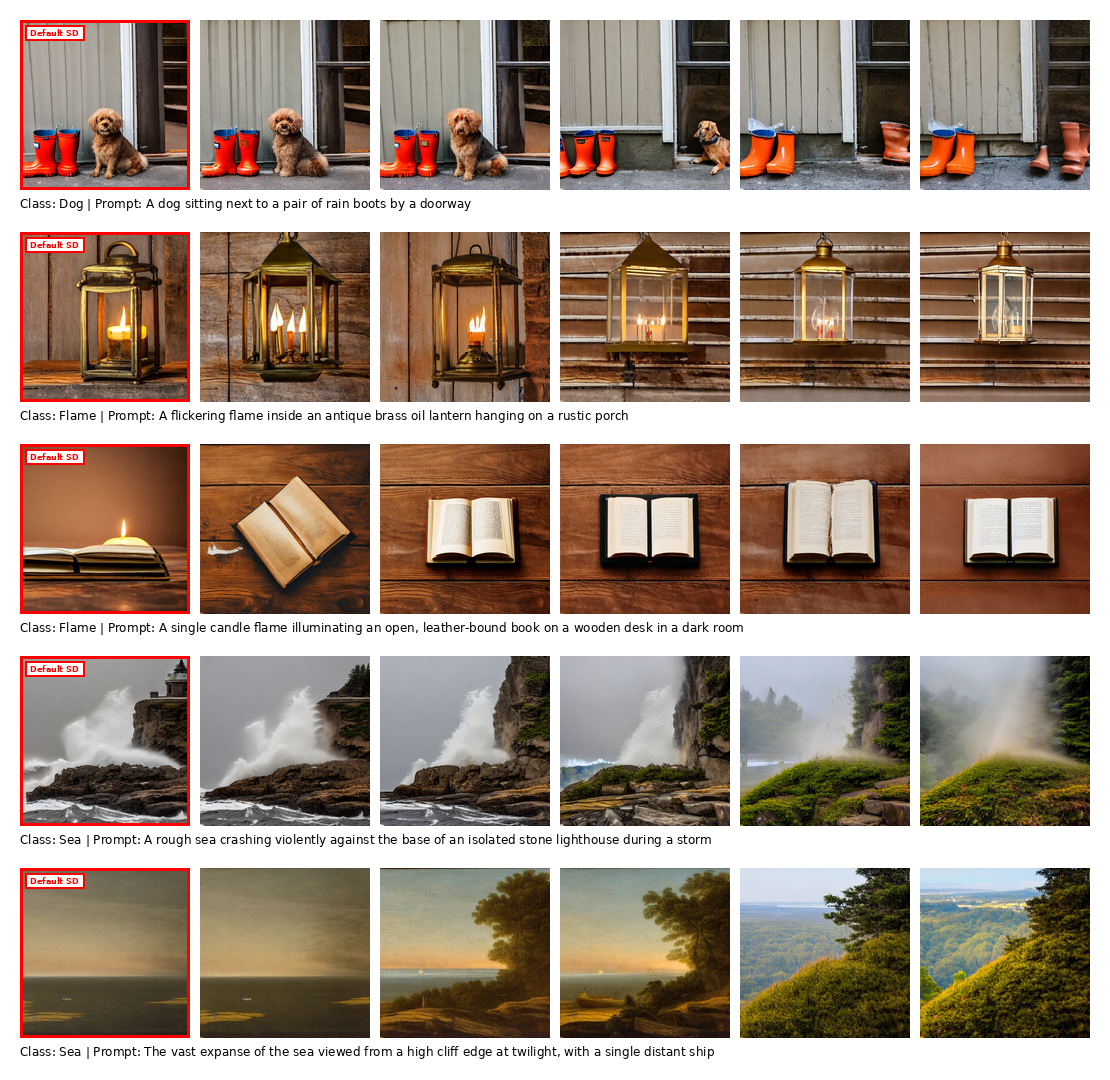}
    \caption{Qualitative progression examples (set 2).}
    \label{fig:progression_appendix_2}
\end{figure}

\clearpage

\subsection{Gradient Variance Analysis}
\label{subsec:grad_var}

\begin{figure}[h]
    \centering
    \includegraphics[width=0.7\textwidth]{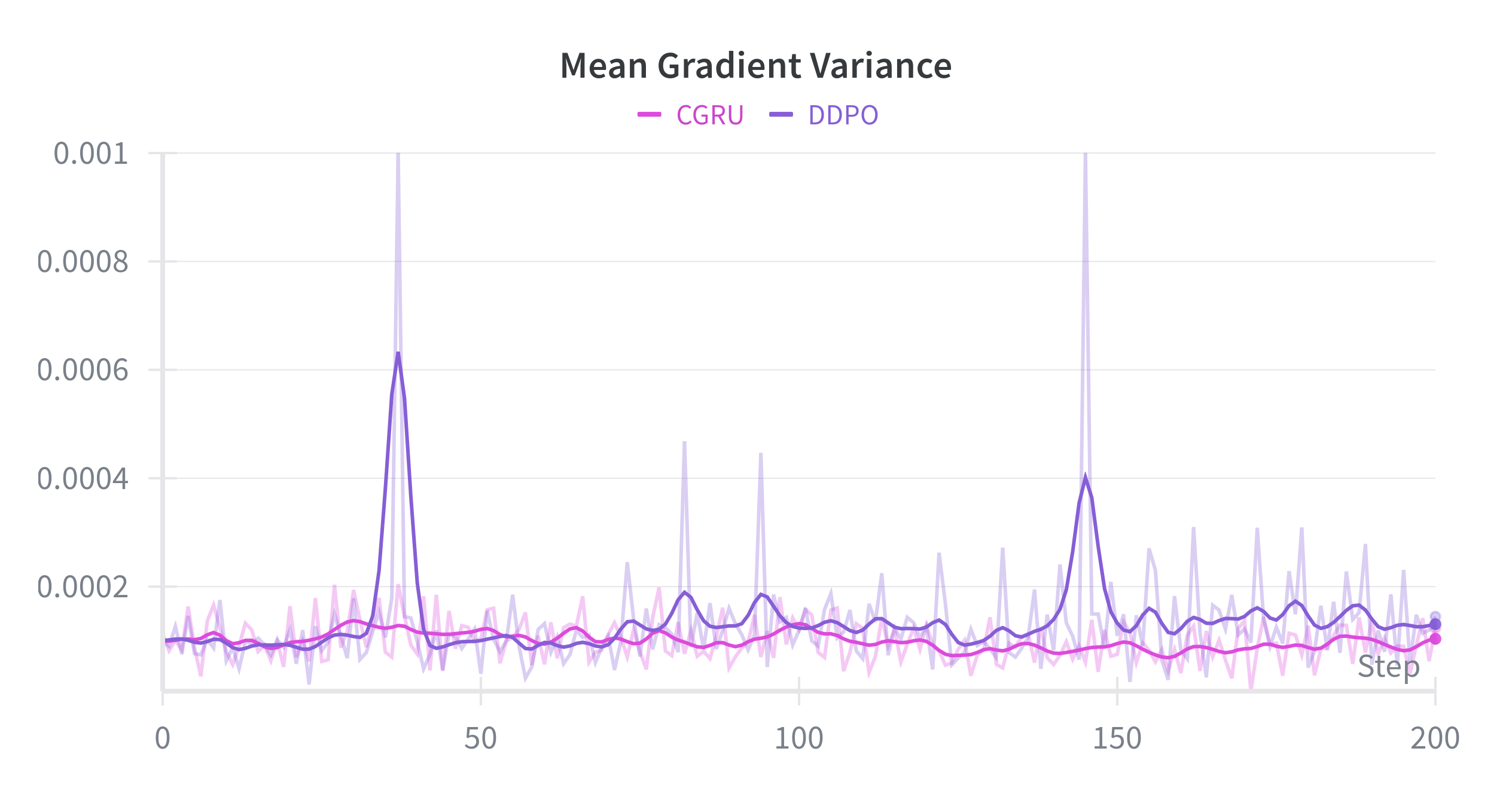}
    \caption{Comparison of mean gradient variance during training for CGRU and DDPO. CGRU exhibits lower and more stable variance, validating the effectiveness of the per-timestep critic for variance reduction.}
    \label{fig:grad_var}
\end{figure}

To empirically validate our theoretical claims regarding stability, we analyzed the variance of the policy gradients during training. Figure~\ref{fig:grad_var} compares the mean gradient variance of CGRU against DDPO on the \textit{Aesthetic score} objective. CGRU demonstrates consistently lower and more stable gradient variance throughout the optimization process, confirming the benefit of the per-timestep critic in reducing the high variance associated with sparse terminal rewards.

\subsection{Importance Sampling Ratio Analysis}
\label{subsec:is_analysis}

\begin{figure}[h]
    \centering
    \includegraphics[width=0.7\textwidth]{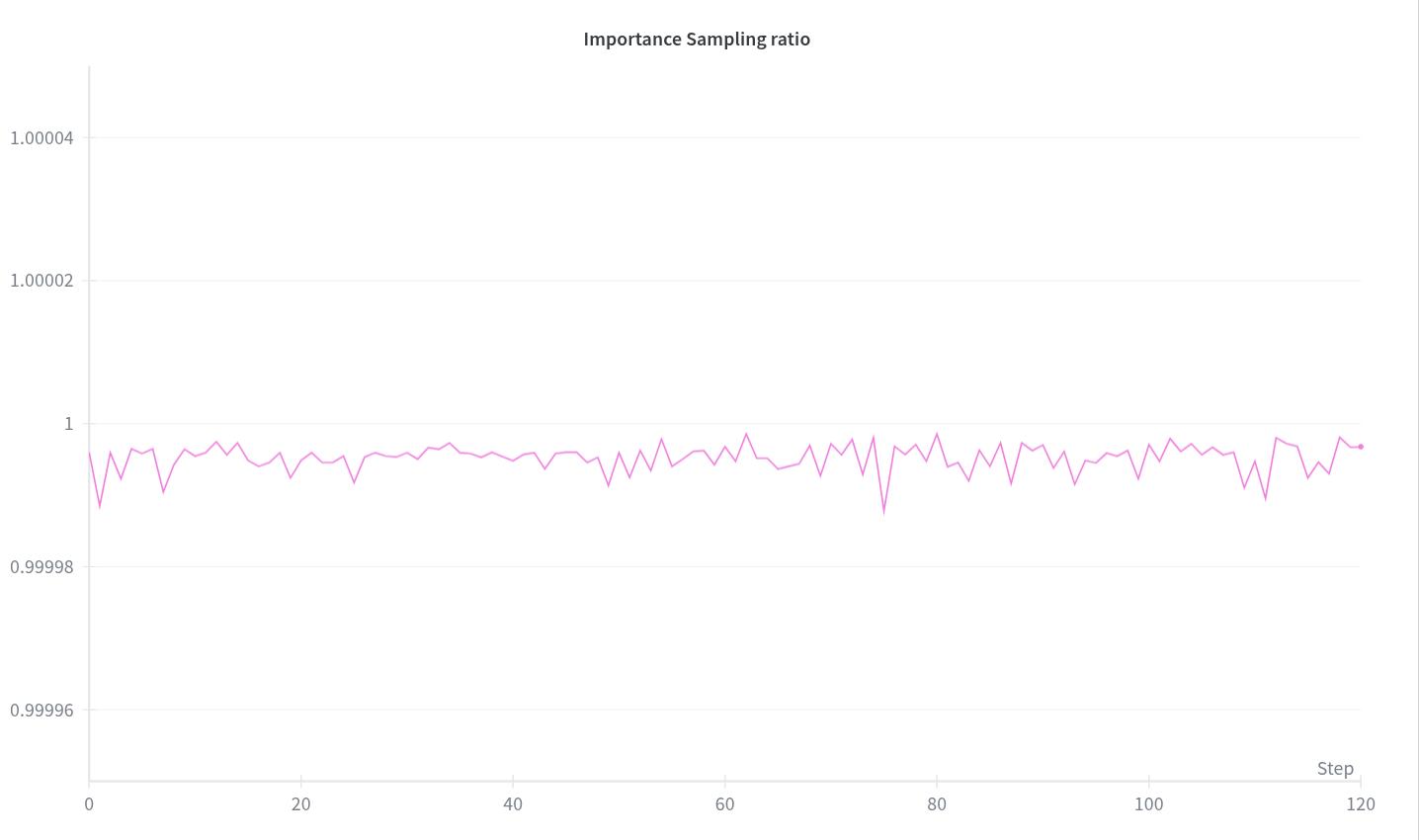}
    \caption{\textbf{Importance-sampling (IS) ratio stability.} IS ratios $\rho_t$ over training (before clipping), aggregated over timesteps and batch. Ratios remain tightly concentrated around $1$, indicating stable near on-policy trajectory reuse.}
    \label{fig:is_mean}
\end{figure}

To assess the stability of trajectory reuse, we monitor the per-timestep importance ratio
$\rho_t = p_{\theta}(x_{t-1} \mid x_t ,c ) / p_{\theta_{\text{old}}}(x_{t-1} \mid x_t , c)$
for each denoising step when updating the policy on trajectories collected by $\theta_{\text{old}}$.
Figure~\ref{fig:is_mean} plots $\rho_t$ (aggregated over the batch and timesteps) across training
\emph{before any clipping}. We observe that the IS ratios remain tightly concentrated around $1$
with small variance and no heavy tails, indicating that policy updates stay in a local regime where
$\theta$ does not drift far from $\theta_{\text{old}}$ between data collection and optimization.
This empirically validates that multi-step reuse is effectively \emph{near on-policy} in our setting
and does not exhibit the instability typically associated with large IS weights; consequently, IS
reweighting provides a safe correction for minor distribution mismatch while preserving stable
training dynamics.

\section{Training Details}
\label{app:training}

\subsection{Hyperparameters}

Table~\ref{tab:hyperparameters} provides the complete hyperparameter configuration used in our concept removal experiments. The configuration was optimized for training on a single H100 GPU, balancing memory efficiency with training stability.

\begin{table}[h]
\centering
\caption{\textbf{Hyperparameter configuration for CGRU training.} All concept removal experiments used these settings unless otherwise specified.}
\label{tab:hyperparameters}
\begin{tabular}{ll}
\toprule
\textbf{Parameter} & \textbf{Value} \\
\midrule
Learning rate & $3 \times 10^{-4}$ \\
Batch size & 2 \\
Gradient accumulation steps & 4 \\
Sampling batch size & 4 \\
Batches per epoch & 4 \\
Number of denoising steps & 50 \\
Training epochs & 100--250 (class-dependent) \\
Gradient max-norm & 1.0 \\
Importance sampling ratio clipping & [$1-10^{-4}, 1+10^{-4}$] \\
\midrule
\textbf{Hardware} & \\
GPU & Single H100 (80GB VRAM) \\
Memory usage & $\sim$35 GB VRAM \\
Training time & $\sim$2--5 hours per run (class-dependent) \\
\bottomrule
\end{tabular}
\end{table}

\subsection{Training Prompts}

Our experiments utilized diverse prompts covering various concepts. Table~\ref{tab:example_prompts} shows representative examples of the prompts used during training, demonstrating the variety of concepts.

The dataset contained 80 prompts for each of 20 classes.

\begin{table}[h]
\centering
\caption{\textbf{Example training prompts.} These prompts represent the diversity of concepts and styles used in our experiments.}
\label{tab:example_prompts}
\begin{tabular}{p{0.9\textwidth}}
\toprule
\textbf{Example Prompts} \\
\midrule
\textit{A bird with a scent of lavender, a walking bloom.} \\
\textit{Rabbit with a mischievous twinkle in its eye.} \\
\textit{A cat painting a self-portrait in a studio.} \\
\textit{A mango tree providing shade in a tropical village.} \\
\textit{Jellyfish resembling a hovering spaceship.} \\
\bottomrule
\end{tabular}
\end{table}

\subsection{Object Classes}

Our concept removal experiments covered 20 distinct object classes, providing comprehensive evaluation across diverse visual concepts. The complete list of classes is provided in Table~\ref{tab:object_classes}.

\begin{table}[h]
\centering
\caption{\textbf{Complete list of object classes used in concept removal experiments.} Each class represents a distinct visual concept that can be targeted for unlearning.}
\label{tab:object_classes}
\begin{tabular}{llll}
\toprule
Architectures & Bears & Birds & Butterfly \\
Cats & Dogs & Fishes & Flame \\
Flowers & Frogs & Horses & Human \\
Jellyfish & Rabbits & Sandwiches & Sea \\
Statues & Towers & Trees & Waterfalls \\
\bottomrule
\end{tabular}
\end{table}

\subsection{Implementation Details}

Our implementation leverages LoRA (Low-Rank Adaptation) \cite{hu2021loralowrankadaptationlarge} for efficient fine-tuning of the diffusion model, significantly reducing memory requirements while maintaining training effectiveness. The gradient accumulation strategy enables effective batch sizes larger than what would fit in GPU memory, while the increased number of denoising steps (50 vs. the typical 20) provides more detailed trajectory information for the critic network.
\clearpage

\section{Training Algorithms}
\label{app:algorithms}

\subsection{Critic Training Algorithm}
\label{alg:critic-training}

\begin{algorithm}[h]
\caption{Critic Training FOR CGRU}
\label{alg:critic}
\begin{algorithmic}[1]
\REQUIRE Dataset of prompts $\mathcal{D}_c = \{c_i\}_{i=1}^N$, diffusion model $p_\theta$, reward function $r(x_0, c)$
\ENSURE Trained critic $V_\phi(x_t, c, t)$

\STATE Initialize critic network $V_\phi$ with random parameters $\phi$
\STATE Initialize empty buffer $\mathcal{B} = \emptyset$

\FOR{epoch $= 1$ to $E_{\text{critic}}$}
    \FOR{each prompt $c \sim \mathcal{D}_c$}
        \STATE Generate trajectory: $x_T \sim \mathcal{N}(0,I)$, $x_{t-1} \sim p_\theta(x_{t-1}|x_t, c)$ FOR $t = T, \ldots, 1$
        \STATE Compute terminal reward: $r_{\text{final}} = r(x_0, c)$
        \STATE Store $(x_t, c, t, r_{\text{final}})$ in $\mathcal{B}$ FOR all $t \in \{1, \ldots, T\}$
    \ENDFOR
    
    \FOR{update $= 1$ to $U_{\text{critic}}$}
        \STATE Sample minibatch $\{(x_t^{(i)}, c^{(i)}, t^{(i)}, r^{(i)})\}_{i=1}^B$ from $\mathcal{B}$
        \STATE Compute critic loss:
        \STATE \quad $\mathcal{L}_{\text{critic}} = \frac{1}{B}\sum_{i=1}^B \|V_\phi(x_t^{(i)}, c^{(i)}, t^{(i)}) - r^{(i)}\|^2$
        \STATE Update critic: $\phi \leftarrow \phi - \alpha_{\text{critic}} \nabla_\phi \mathcal{L}_{\text{critic}}$
    \ENDFOR
\ENDFOR

\end{algorithmic}
\end{algorithm}

\subsection{Policy Training Algorithm}
\label{alg:policy-training}

\begin{algorithm}[H]
\caption{Policy Training with CGRU}
\label{alg:policy}
\begin{algorithmic}[1]
\REQUIRE Dataset of prompts $\mathcal{D}_c = \{c_i\}_{i=1}^N$, critic $V_\phi$ (warm-started), diffusion model $p_\theta$
\ENSURE Updated diffusion model $p_{\theta'}$

\STATE Initialize policy parameters $\theta_{\text{old}} \leftarrow \theta$
\STATE Initialize empty trajectory buffer $\mathcal{T} = \emptyset$
\STATE Initialize empty critic buffer $\mathcal{B} = \emptyset$

\FOR{iteration $= 1$ to $I$}
    \STATE Set behavior policy parameters $\theta_{\text{old}} \leftarrow \theta$
    \STATE Clear trajectory buffer $\mathcal{T} \leftarrow \emptyset$
    \STATE Clear critic buffer $\mathcal{B} \leftarrow \emptyset$
    \FOR{each prompt $c \sim \mathcal{D}_c$}
        \STATE Generate trajectory: $x_T \sim \mathcal{N}(0,I)$, $x_{t-1} \sim p_\theta(x_{t-1}|x_t, c)$ FOR $t = T, \ldots, 1$
        \STATE Compute terminal reward: $r_{\text{final}} = r(x_0, c)$
        \STATE Compute advantages: $A_t = r_{\text{final}} - V_\phi(x_t, c, t)$ FOR all $t$
        \STATE Store transitions $\tau = \{(x_t, x_{t-1}, c, t, A_t)\}_{t=1}^T$ in $\mathcal{T}$
        \STATE Store $(x_t, c, t, r_{\text{final}})$ in $\mathcal{B}$ FOR all $t \in \{1, \ldots, T\}$
    \ENDFOR

    \FOR{update $= 1$ to $U_{\text{critic}}^{\text{online}}$}
        \STATE Sample minibatch $\{(x_t^{(i)}, c^{(i)}, t^{(i)}, r^{(i)})\}_{i=1}^B$ from $\mathcal{B}$
        \STATE Compute critic loss:
        \STATE \quad $\mathcal{L}_{\text{critic}} = \frac{1}{B}\sum_{i=1}^B \|V_\phi(x_t^{(i)}, c^{(i)}, t^{(i)}) - r^{(i)}\|^2$
        \STATE Update critic: $\phi \leftarrow \phi - \alpha_{\text{critic}} \nabla_\phi \mathcal{L}_{\text{critic}}$
    \ENDFOR
    
    \FOR{update $= 1$ to $U_{\text{policy}}$}
        \STATE Sample minibatch $\{(x_t^{(i)}, x_{t-1}^{(i)}, c^{(i)}, t^{(i)}, A_t^{(i)})\}_{i=1}^B$ from $\mathcal{T}$
        \STATE Compute importance weights:
        \STATE \quad $w^{(i)} = \frac{p_\theta(x_{t-1}^{(i)}|x_t^{(i)}, c^{(i)})}{p_{\theta_{\text{old}}}(x_{t-1}^{(i)}|x_t^{(i)}, c^{(i)})}$
        \STATE Compute policy update:
        \STATE \quad $\nabla_\theta\mathcal{L}_{\text{policy}} = -\frac{1}{B}\sum_{i=1}^B w^{(i)} \nabla_\theta \log p_\theta(x_{t-1}^{(i)}|x_t^{(i)}, c^{(i)}) \cdot A_t^{(i)}$
        \STATE Update policy: $\theta \leftarrow \theta - \alpha_{\text{policy}} \nabla_\theta \mathcal{L}_{\text{policy}}$
    \ENDFOR
\ENDFOR

\end{algorithmic}
\end{algorithm}

\subsection{Complete CGRU Training Pipeline}
\label{alg:complete-pipeline}

\begin{algorithm}[H]
\caption{Complete CGRU Training Pipeline}
\label{alg:complete}
\begin{algorithmic}[1]
\REQUIRE Concept to unlearn $\mathcal{C}$, base diffusion model $p_\theta^{\text{base}}$, prompt datasets $\mathcal{D}_c$
\ENSURE Unlearned diffusion model $p_\theta^{\text{unlearned}}$

\STATE \textbf{Step 1: Define Reward Function}
\STATE Design $r(x_0, c)$ to penalize concept $\mathcal{C}$ while preserving utility
\STATE \quad (e.g., using CLIP-based similarity or classifier outputs)

\STATE \textbf{Step 2: Warm-start Critic}
\STATE $V_\phi \leftarrow$ \textbf{TrainCritic}($p_\theta^{\text{base}}$, $r$, $\mathcal{D}_c$)

\STATE \textbf{Step 3: Train Policy (with online critic updates)}
\STATE $p_\theta^{\text{unlearned}} \leftarrow$ \textbf{TrainPolicy}($p_\theta^{\text{base}}$, $V_\phi$, $\mathcal{D}_c$)

\end{algorithmic}
\end{algorithm}

\section{Feature-wise Linear Modulation (FiLM) Layers}
\label{app:film}

Feature-wise Linear Modulation (FiLM) layers~\cite{perez2017filmvisualreasoninggeneral} are a conditioning mechanism that allows neural networks to adapt their behavior based on external conditioning information. In our critic architecture, FiLM layers enable the network to modulate its intermediate representations based on the timestep information.

\subsection{FiLM Layer Formulation}

A FiLM layer takes two inputs: (1) the feature map $x$ from the previous layer, and (2) a conditioning vector $\gamma$ that encodes the timestep information. The layer applies an affine transformation to each feature channel:

\begin{equation}
\text{FiLM}(x, \gamma) = \gamma_{\text{scale}} \odot x + \gamma_{\text{shift}}
\end{equation}

where $\gamma_{\text{scale}}$ and $\gamma_{\text{shift}}$ are learned parameters derived from the conditioning vector $\gamma$, and $\odot$ denotes element-wise multiplication.

\subsection{Integration in Critic Architecture}

In our critic network, the timestep $t$ is first encoded using sinusoidal embeddings to create a dense representation. This timestep embedding is then processed through a small MLP to generate the conditioning parameters $\gamma_{\text{scale}}$ and $\gamma_{\text{shift}}$ for each FiLM layer. The FiLM layers are strategically placed throughout the network to allow timestep-dependent modulation of feature representations.

This design enables the critic to learn different value estimation strategies for different timesteps in the denoising process, which is crucial for providing accurate baselines throughout the trajectory. The FiLM conditioning allows the network to adapt its internal representations based on whether it is processing early (noisy) or late (clean) stages of the denoising process.

\subsection{Ablation Study}

\begin{figure}[h]
    \centering
    \includegraphics[width=0.8\textwidth]{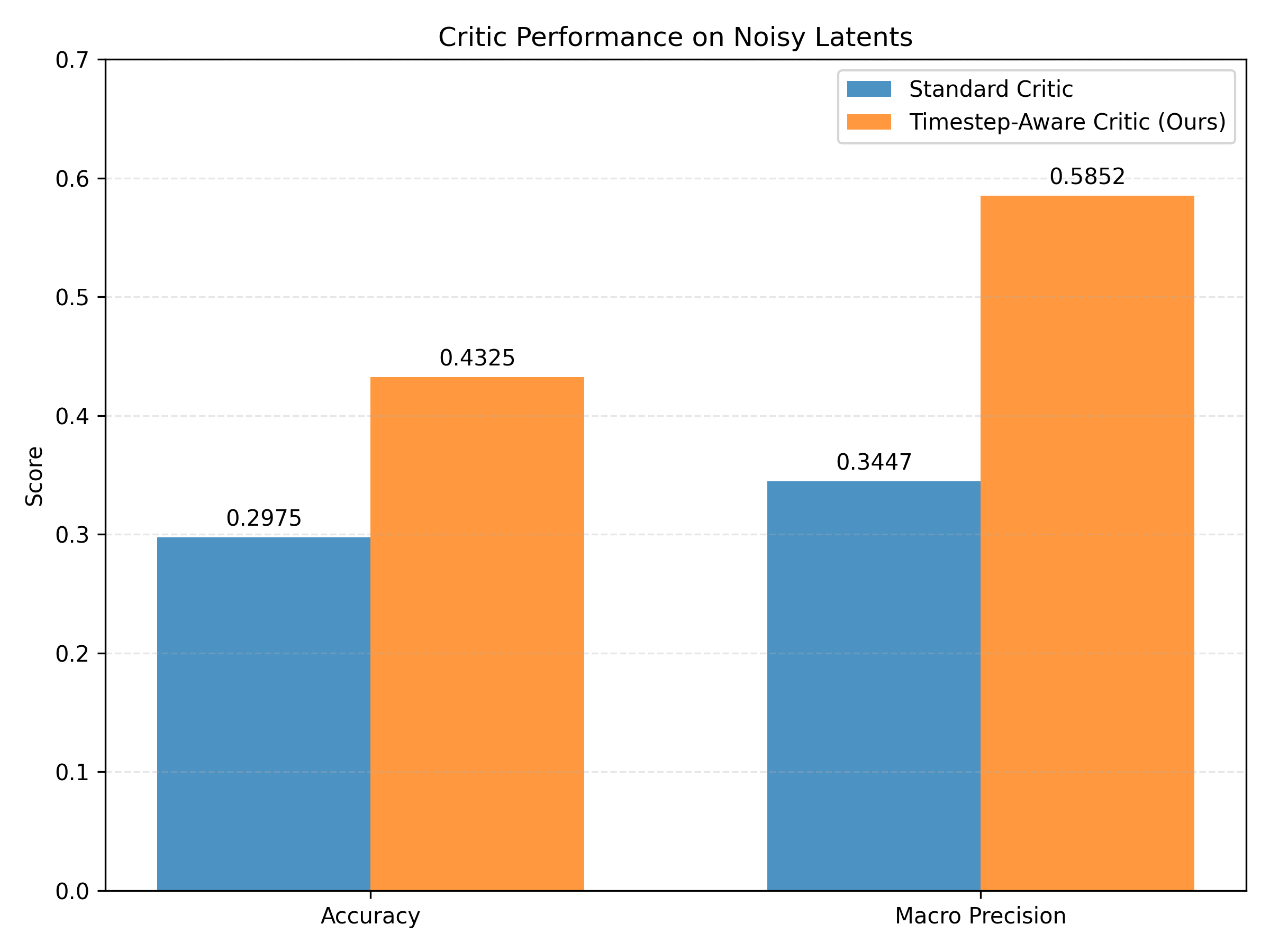}
    \caption{Ablation study comparing the performance of a standard critic vs. our timestep-aware critic on noisy latents. The timestep-aware critic achieves substantially higher accuracy (43.25\% vs. 29.75\%) and macro precision (58.52\% vs. 34.47\%).}
    \label{fig:ablation_metrics}
\end{figure}

Figure~\ref{fig:ablation_metrics} compares our timestep-aware critic against a standard CLIP-based critic without FiLM conditioning. Both models are evaluated on their ability to infer the concept signal used to define the terminal reward from intermediate noisy latents. Since the reward is computed from a 20-class concept classifier (via the target-class probability), we report \emph{classification} metrics (accuracy and macro precision) on the induced discrete class predictions (argmax over the 20-way outputs). Test samples are drawn uniformly across timesteps to cover the full noise spectrum.

\section{Evaluation metrics}
\label{app:evalmetr}
\subsection{Fr\'echet Inception Distance (FID) Computation}

For distribution-level image quality, we report the Fr\'echet Inception Distance (FID). Specifically:

\begin{itemize}
    \item \textbf{Backbone:} Inception-V3, pool3 activations (2048-D).
    
    \item \textbf{Computation:} Let $\mu_r,\Sigma_r$ be the mean and covariance of reference features, and $\mu_g,\Sigma_g$ those of generated features. FID is
    \[
        \mathrm{FID}(\mathcal{R},\mathcal{G}) \;=\; \|\mu_r - \mu_g\|_2^2 \;+\; \mathrm{Tr}\!\big(\Sigma_r + \Sigma_g - 2(\Sigma_r \Sigma_g)^{1/2}\big).
    \]
    For numerical stability we add a small diagonal term ($10^{-6}I$) to covariances prior to the matrix square root.
\end{itemize}
\subsection{CLIP Score Computation}
\label{app:clipscore}

For image--text alignment evaluation, we report the CLIP score as a cosine similarity between image and text embeddings, following standard practice. Specifically:

\begin{itemize}
    \item \textbf{Model:} We use the OpenAI CLIP ViT-L/14@336px encoder.
    \item \textbf{Computation:} Given a generated image $I$ and its corresponding text prompt $T$, we obtain the normalized image embedding $f(I)$ and text embedding $g(T)$. The CLIP score is then
    \[
        \text{CLIP}(I, T) = \frac{f(I) \cdot g(T)}{\| f(I) \| \, \| g(T) \|} \in [0, 1].
    \]
    \textbf{Note:} All CLIP scores reported in this paper are multiplied by 100 for readability.

    \item \textbf{Aggregation:} For each model, we report the mean CLIP score over the entire evaluation set (30k prompts from MJHQ).
    \item \textbf{Variance reporting:} We additionally compute the standard deviation across random seeds.
\end{itemize}

\section{Large Language Model Usage}
\label{app:llm_usage}

We used Large Language Models (LLMs) to aid and polish the writing of this paper. The LLMs were employed for:

\begin{itemize}
\item \textbf{Writing assistance}: Helping with sentence structure, clarity, and flow
\item \textbf{Language polishing}: Improving grammar, style, and academic tone
\item \textbf{Content organization}: Assisting with logical flow and section transitions
\item \textbf{Technical writing}: Ensuring consistent terminology and precise mathematical descriptions
\end{itemize}

All technical content, experimental results, mathematical formulations, and scientific claims remain entirely our own. The LLMs were used solely as writing tools to improve the presentation and readability of our research, following standard academic practices for manuscript preparation.

\subsection{Safety-Aware Instruction for NSFW Prompt Generation}
\label{app:llava_safety_prompt}

For the NSFW unlearning experiment (Section~\ref{subsec:nsfw}), we re-caption training images into text prompts using LLaVA-1.5-7B with the following safety-aware instruction:
\begin{verbatim}
CAPTION_USER_PROMPT = (
    "Describe this image in one concise sentence (15-25 words): "
    "main subject, pose/action, setting. "
    "Include what NSFW or suggestive elements are present in detail: 
    nudity, revealing clothing, poses, etc."
    "Make it sound like a stable diffusion prompt."
)
\end{verbatim}


\end{document}